\documentclass{article}

\usepackage{threeparttable}
\usepackage[table]{xcolor} 
\usepackage{enumitem}
\definecolor{highlight}{gray}{0.9}
\usepackage[preprint]{neurips_2026}

\usepackage{float}

\usepackage[utf8]{inputenc} 
\usepackage[T1]{fontenc}    
\usepackage{hyperref}       
\usepackage{url}            
\usepackage{booktabs}       
\usepackage{amsfonts}       
\usepackage{nicefrac}       
\usepackage{microtype}      
\usepackage{xcolor}         

\usepackage{graphicx}
\usepackage{amsmath}
\usepackage{amsthm}
\usepackage{amssymb}
\usepackage[ruled,vlined]{algorithm2e}
\usepackage{multirow}
\usepackage[most]{tcolorbox}

\newcommand{\cL}{\mathcal{L}}
\newcommand{\NS}{\texttt{NS5}}

\newcommand{\msgn}{\texttt{msgn}}
\DeclareMathOperator*{\argmin}{arg\,min}
\usepackage{caption}
\allowdisplaybreaks
\newtheorem{remark}{Remark}

\title{Muon-OGD: Muon-based Spectral Orthogonal Gradient Projection for LLM Continual Learning }

\author{%
  Binghang Lu\thanks{Corresponding authors: \texttt{lu895@purdue.edu}, \texttt{runyuzha@mit.edu}, \texttt{xiaominli@g.harvard.edu}.} \\
  Purdue University
  \And
  Zheyuan Deng\thanks{Co-second authors.} \\
  Brown University
  \And
  Runyu Zhang\footnotemark[2]\footnotemark[1] \\
  Massachusetts Institute of Technology
  \AND
  Bing Hu \\
  Independent Researcher
  \And
  Yunhan Zhao \\
  University of California, Irvine
  \And
  Yuan Tian \\
  Independent Researcher
  \AND
  Changhong Mou \\
  Utah State University 
  \And
  Guang Lin \\
  Purdue University
  \And
  Xiaomin Li\footnotemark[1] \\
  Harvard University
}

\begin{document}

\maketitle

\begin{abstract}
 A central challenge in continual learning for large language models (LLMs) is catastrophic forgetting, where adapting to new tasks can substantially degrade performance on previously learned ones. Existing projection-based methods mitigate such interference by restricting parameter updates to subspaces that are orthogonal to directions associated with past tasks. However, these methods are typically formulated under Euclidean parameter geometry, with update magnitudes and projections governed by the Frobenius norm. The recent empirical success of the Muon optimizer, which applies orthogonalized matrix updates and admits a spectral-norm interpretation, suggests that Frobenius geometry may not be the most effective choice for matrix-valued LLM parameters. Motivated by this observation, we propose Muon-OGD, a spectral-norm-aware continual learning framework that integrates Muon-style operator-norm geometry with orthogonal projection constraints. Our method formulates each update as a spectral-norm-constrained optimization problem with linear non-interference constraints, and solves it efficiently through dual iterations and Newton--Schulz matrix-sign approximations. By applying orthogonalized momentum updates that avoid protected directions associated with prior tasks, Muon-OGD aims to improve the stability--plasticity trade-off in sequential LLM adaptation. We evaluate the proposed method on standard continual learning benchmarks, TRACE, and domain-specific Coding--Math--Medical curricula using both encoder--decoder and decoder-only architectures. Empirically, Muon-OGD consistently improves over sequential fine-tuning and competitive orthogonal-gradient baselines, while remaining computationally scalable. These results suggest that spectral-norm-aware update geometry provides a practical and effective alternative to Frobenius-norm projection for continual learning in LLMs.
\end{abstract}
\vspace{-8pt}
\section{Introduction}
\label{sec:intro}
\vspace{-5pt}

Large-scale neural networks, including Large Language Models (LLMs), are increasingly deployed in settings that require sequential adaptation across diverse domains. A central challenge in such continual or post-training regimes is catastrophic forgetting \cite{kirkpatrick2017overcoming,goodfellow2013empirical}: when models are fine-tuned on new tasks, their performance on previously learned tasks can degrade rapidly. This phenomenon has been extensively documented in the literature on neural networks and, more recently, in large-scale models such as large language models (LLMs). 

Early empirical and theoretical studies demonstrate that standard gradient-based training does not preserve previously acquired knowledge in the absence of explicit mechanisms \cite{kirkpatrick2017overcoming, shin2017continual, yoon2018lifelong}. To address this issue, the field of continual learning (also known as lifelong learning) studies how to train models on a sequence of tasks while retaining performance on past tasks. Foundational works establish core problem formulations and evaluation protocols for this setting \cite{li2017learning, kirkpatrick2017overcoming, lopez2017gradient}. Formally, continual learning considers a sequence of tasks, where the learner observes data from each task in turn and must update its parameters using only \emph{limited or no access to past data}, while maintaining strong performance across all previously encountered tasks. The central objective is to balance \emph{plasticity} (the ability to learn new tasks) and \emph{stability} (the ability to retain prior knowledge), a tradeoff that lies at the heart of continual learning research.
\begin{figure}[htbp]
    \centering
    \includegraphics[width=0.9\textwidth]{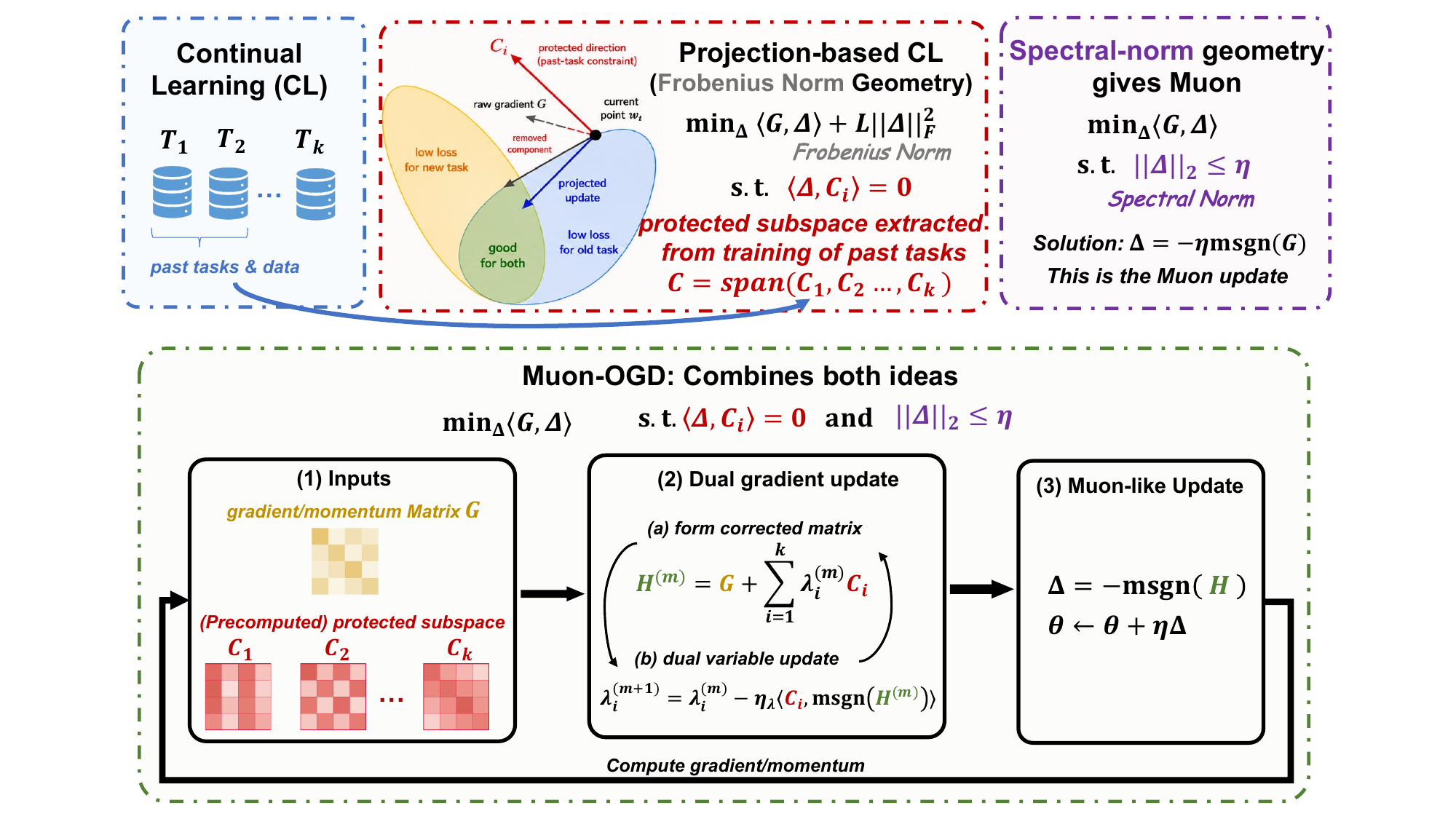} 
    \caption{Overview of our Muon-OGD Continual Fine-tuning method: (Top Left) Traditional projection-based continual learning methods operate under Frobenius-norm geometry, mitigating catastrophic forgetting by projecting the raw gradient $G$ away from protected past-task directions $\{C_i\}$. (Top Right) The Muon optimizer performs steepest descent under spectral-norm geometry, enforcing an operator-norm bound $||\Delta||_2 \le \eta$. (Bottom) Muon-OGD bridges these paradigms by formulating parameter updates as a spectral-norm-aware constrained optimization problem. The optimal update is solved efficiently via dual gradient updates, which iteratively form a corrected momentum matrix $H$ before applying the final Muon-style matrix-sign ($\mathrm{msg}$) update.}
    \label{fig:muon_ogd_overview}
\end{figure}

A prominent class of continual learning methods mitigates forgetting by explicitly constraining parameter updates to avoid interference with previously learned tasks. Examples include GEM\cite{lopez2017gradient}, A-GEM \cite{chaudhry2018efficient}, Orthogonal Gradient Descent (OGD)\cite{farajtabar2020orthogonal}, O-LoRA \cite{xiong2026oplora} and Sculpting Subspaces \cite{nayak2025sculpting}. At a high level, these methods identify directions associated with past tasks and restrict updates so as not to alter them, thereby preserving previously acquired knowledge. This idea can be formalized as a constrained optimization problem at each update step. Given a gradient or momentum vector $G$ and a collection of constraint directions $\{C_i\}_{i=1}^k$, the update $\Delta$ is obtained by solving
\begin{equation}\label{eq:OGD-intro}
\begin{split}
\textstyle \min_{\Delta\in \mathbb{R}^{m\times n}}~~ \langle G, \Delta\rangle + \frac{L}{2}\|\Delta\|_F^2 \qquad 
\text{s.t.}~~  \langle C_i, \Delta\rangle = 0, ~~ i=1,\dots,k,
\end{split}
\end{equation}
which seeks a descent direction that improves the current task while remaining orthogonal to directions that would degrade performance on previous tasks. Such projection-based updates have been shown to effectively balance adaptation and retention in practice.

It is worth noting that the formulation in \eqref{eq:OGD-intro} relies on the Frobenius norm $\|\Delta\|_F^2$ to measure update magnitude and define orthogonality, implicitly treating all directions in parameter space uniformly. However, recent advances in optimization for deep learning suggest that, for matrix-shaped parameters, controlling the operator (spectral) norm of updates can be more effective for maintaining good conditioning and stable training dynamics. In particular, \cite{jordan_muon} proposes the Muon optimizer for neutral network, especially LLM, training. Its core idea is that, for matrix-shaped parameters, to replace the gradient update (e.g., momentum or EMA of gradients) with an \emph{orthogonalized} matrix direction obtained via an approximate singular value decomposition (SVD). To save computation, rather than computing an explicit SVD, Muon employs a small number of Newton--Schulz iterations to efficiently approximate, making it practical at scale. This simple updates has shown surprisingly efficient convergence \cite{liu2025muon,jordan_muon} that even outperform the state of the art optimizer such as AdamW \cite{loshchilov2017decoupled}. Given its success, there are many works trying to understand why it is so, and one common view point  is that the muon update can be  interpreted as solving
\begin{equation}\label{eq:muon-intro}
\begin{split}
\textstyle \min_{\Delta\in\mathbb{R}^{m\times n}}~~  \langle G, \Delta\rangle \qquad
\text{s.t.}~~  \|\Delta\|_2 \le \eta ,
\end{split}
\end{equation}
which corresponds to a steepest descent step under a spectral-norm constraint. And there are many works suggesting that this spectral-norm has better property than the Frobenius norm \cite{shen2025convergence,su2025isotropic,davis2025spectral}, suggesting that the choice of norm may play a critical role in shaping update behavior.

The above viewpoints reveal a fundamental mismatch between the geometry underlying modern optimization methods and that used in projection-based continual learning. While the update in \eqref{eq:OGD-intro} enforces stability through Frobenius-norm-based orthogonality constraints, the Muon formulation in \eqref{eq:muon-intro} selects descent directions according to spectral-norm geometry. As a result, existing continual learning mechanisms are not naturally aligned with the geometry of updates induced by optimizers such as Muon, particularly for matrix-valued parameters.

Our core idea is to bridge this gap by integrating spectral-norm-aware optimization with projection-based continual learning constraints as shown in Figure~\ref{fig:muon_ogd_overview}. Specifically, we propose to compute updates by solving a constrained optimization problem that enforces both (i) orthogonality to protected directions associated with past tasks, and (ii) a spectral-norm constraint that controls the operator magnitude of updates,
which can be viewed as a spectral-norm-constrained analogue of \eqref{eq:OGD-intro}. We refer to the resulting method as \emph{Muon-OGD}, which combines the stability guarantees of projection-based methods with the conditioning benefits of spectral-norm-aware updates. In the next section, we show how this problem can be solved efficiently via a dual formulation, leading to a practical algorithm.

\paragraph{Our Contributions} 
Our contributions are threefold.
First, we formulate projection-based continual learning under a spectral-norm geometry. Unlike existing methods that rely on Frobenius-norm regularization or projection, our formulation constrains the operator norm of the update while enforcing orthogonality to protected past-task directions. This provides a principled way to align continual learning constraints with the geometry of modern spectral-norm-aware optimizers such as Muon. Second, we derive a practical Muon-OGD update rule for the resulting constrained optimization problem. The proposed method combines past-task orthogonality constraints with Muon-style matrix-sign updates, and solves the constrained update efficiently through a dual formulation. This leads to an optimizer-compatible algorithm that preserves the stability benefits of projection-based continual learning while retaining the conditioning advantages of spectral-norm-aware updates. Third, we instantiate the framework for LLM continual fine-tuning by adapting the constraint-direction construction and redesigning the update rule under our spectral-norm-constrained formulation. Extensive experiments on standard continual learning benchmarks, the 15-Task CL Benchmark, and TRACE demonstrate that Muon-OGD consistently improves the plasticity--stability tradeoff over Frobenius-norm-based counterparts, achieving stronger retention and overall continual learning performance.

\vspace{-5pt}
\section{Related Work}
\label{sec:related_work}
\vspace{-5pt}
\paragraph{Continual Learning in Large Language Models.}
Continual learning studies how to adapt models to sequential tasks while retaining prior knowledge~\cite{parisi2019continual,de2021continual,wang2024comprehensive}. Classical methods include regularization-based approaches~\cite{kirkpatrick2017overcoming,zenke2017continual,aljundi2018memory,li2017learning}, replay-based approaches~\cite{lopez2017gradient,chaudhry2018efficient}, and prompt- or parameter-efficient methods~\cite{wang2022learning,wang2022dualprompt,razdaibiedina2023progressive,qin2021lfpt5,hu2022lora}. In LLMs, sequential instruction tuning can degrade broad capabilities such as reasoning and instruction following, motivating benchmarks such as TRACE and long multi-task streams~\cite{wang2023trace,razdaibiedina2023progressive,wang2023orthogonal,nayak2025sculpting}.
\paragraph{Projection-based Continual Learning.}
Projection-based methods reduce forgetting by constraining update directions to avoid interference with previous tasks. GEM and A-GEM impose episodic-memory gradient constraints~\cite{lopez2017gradient,chaudhry2018efficient}, while OGD projects new gradients away from subspaces important for prior tasks~\cite{farajtabar2020orthogonal}. Recent LLM methods extend this principle to low-rank or full-parameter adaptation, including O-LoRA, Sculpting Subspaces, and OSFT~\cite{wang2023orthogonal,nayak2025sculpting}. These methods demonstrate the effectiveness of orthogonal subspace constraints, but they are mostly formulated under Euclidean or SVD-based geometry. Our method instead imposes such constraints under a spectral-norm-aware update geometry.
\paragraph{Spectral-Norm-Aware Optimization and Muon.}
Muon is a recent optimizer for matrix-valued neural-network parameters that orthogonalizes momentum updates using Newton--Schulz iterations~\cite{jordan_muon}. Empirical and theoretical studies suggest that Muon scales well to LLM training and can be interpreted through operator-norm or spectral-norm geometry~\cite{liu2025muon,bernstein2025modular,chen2025muon}, in contrast to coordinate-wise optimizers such as AdamW~\cite{loshchilov2017decoupled}. Muon-inspired ideas have also been explored for low-rank adaptation. Our work connects this spectral-norm-aware optimization perspective with projection-based continual learning.

\vspace{-5pt}
\section{Preliminaries}
\label{sec:preliminary}
\vspace{-5pt}
We briefly review the two ingredients underlying our method: projection-based continual learning and Muon-style spectral-norm-aware optimization.

\vspace{-1pt}
\subsection{Projection-Based Continual Learning}
Projection-based continual learning methods mitigate catastrophic forgetting by restricting new updates to directions that minimally interfere with previous tasks. Representative examples include Orthogonal Gradient Descent (OGD)~\cite{farajtabar20orthogonal}, which projects gradients away from past-task directions, and episodic-memory methods such as GEM~\cite{lopez2017gradient} and A-GEM~\cite{chaudhry2018efficient}, which enforce non-increasing loss on stored previous-task samples. Recent LLM methods further extend this principle to LoRA or full fine-tuning by constraining updates away from subspaces associated with pretrained or previous-task representations~\cite{zhang2025dynamicorthogonalcontinualfinetuning,tekmen2025gem}.

A common formulation is a local constrained quadratic problem. Given a raw gradient matrix $G$ and protected directions $\{C_i\}_{i=1}^k$, the update $\Delta$ is obtained by
\begin{equation}\label{eq:OGD}
\begin{split}
\textstyle \min_{\Delta\in \mathbb{R}^{m\times n}}~~
\langle G, \Delta\rangle + \frac{L}{2}\|\Delta\|_F^2
\qquad 
\text{s.t.}~~ 
\langle C_i, \Delta\rangle = 0, ~~ i=1,\dots,k,
\end{split}
\end{equation}
where $\|\cdot\|_F$ is the Frobenius norm and $\langle A,B\rangle=\mathrm{tr}(A^\top B)$ is the Frobenius inner product. Let
$\mathcal{C}:=\mathrm{span}\{C_1,\dots,C_k\}$. The solution is
\begin{align*}
   \textstyle \Delta^\star
    = -\frac{1}{L}\mathrm{Proj}_{\mathcal{C}^\perp}(G)
    = -\frac{1}{L}\bigl(G-\mathrm{Proj}_{\mathcal{C}}(G)\bigr).
\end{align*}

Thus, the update removes the component of $G$ lying in the protected subspace. This gives a simple constrained-optimization view of OGD-style methods, but it is inherently based on Euclidean/Frobenius geometry.

\vspace{-5pt}
\subsection{The Muon Optimizer and Spectral-Norm Geometry}
\vspace{-5pt}
Muon (Momentum Orthogonalized by Newton--Schulz) is a recently proposed optimizer (cf. \cite{jordan_muon,liu2025muon}), which has gained more and more attention due to its outstanding efficient and performance. It is designed primarily for matrix-shaped (2D) parameters in neural networks. Its core idea is to replace the raw first-moment update $G$ (e.g., momentum) with an \emph{orthonormalized} matrix direction obtained via an approximate singular value decomposition, i.e. if $G = U \Sigma V^\top$ is the singular value decomposition, then the optimal update is:
\vspace{-5pt}
\begin{align}\label{eq:matrix-orthogonalization}
\qquad\Delta^\star = -\eta U V^\top.
\end{align}
However, in practice, computing the exact SVD for very large weight matrices is computationally inefficient. Thus, Muon approximates the matrix orthonormalization \eqref{eq:matrix-orthogonalization} using a small number of Newton--Schulz iterations (cf. \cite{higham1995matrix}):
\vspace{-5pt}
\begin{align*}
    \qquad\qquad UV^\top =: \msgn(G) \approx \NS(G)
\end{align*}
where $\msgn:= UV^\top$ denotes matrix sign of $G$ and the Newton-Schulz update $\NS$ is defined iteratively as:
\vspace{-10pt}
\begin{align*}
   \textstyle \NS(X) := f\circ f\circ f\circ f\circ f(X), \quad f(X) = \frac{3}{2}X - \frac{1}{2}XX^\top X.
\end{align*}

From an optimization perspective, Muon's update can be interpreted as solving, for each matrix block, a linear minimization problem over the spectral-norm unit ball (\cite{bernsteinmodular,bernstein2025deriving}). Given a gradient-like matrix $G\in\mathbb{R}^{m\times n}$, consider:
\vspace{-5pt}
\begin{equation}\label{eq:muon}
\begin{split}
\qquad \textstyle\min_{\Delta\in\mathbb{R}^{m\times n}}~~ \langle G, \Delta\rangle \qquad 
 \text{s.t.}~~ \|\Delta\|_2 \le \eta .
\end{split}
\end{equation}
where $\|\cdot\|_2$ is the induced \emph{matrix $2$-norm (spectral norm)}. One can show that the optimal solution is given by exactly \eqref{eq:matrix-orthogonalization}. Under this lens, Muon can be understood as performing steepest descent under the induced 2-norm geometry. 

\vspace{-5pt}
\section{Methodology}
\label{sec:methodology}
\vspace{-5pt}
The preliminaries above reveal a natural way to combine projection-based continual learning with Muon-style optimization. Projection-based methods protect past-task knowledge by enforcing orthogonality to constraint directions, whereas Muon chooses matrix updates through a spectral-norm-constrained steepest descent step. We therefore formulate continual learning updates directly under spectral-norm geometry, imposing orthogonality constraints from past tasks while selecting the best current-task descent direction within the spectral-norm ball. This leads to a Muon-compatible analogue of classical OGD-style updates.
In particular, we propose to formulate continual learning updates for matrix parameters as the solution of a spectral-norm–aware constrained optimization problem of the form
\vspace{-5pt}
\begin{equation}\label{eq:muon-OGD}
\begin{split}
&\textstyle \min_{\Delta \in \mathbb{R}^{m\times n}} \quad 
\langle G, \Delta\rangle  \\
\text{s.t.}\quad 
& \textstyle \|\Delta\|_2 \le \eta,\quad \langle C_i, \Delta\rangle = 0, \quad i = 1, \dots, k,
\end{split}
\end{equation}
where $G$ denotes the current task gradient (or momentum) and $\{C_i\}_{i=1}^k$ encode constraint directions associated with previously learned tasks (e.g., gradients on episodic memory or protected subspaces). This formulation seeks the descent direction that is optimal under the spectral-norm geometry induced by Muon, while remaining orthogonal to all directions that would interfere with past tasks. By replacing the traditional Frobenius-norm regularization with an induced $2$-norm penalty, this approach naturally blends the operator-norm–aware geometry of Muon with the projection-based stability guarantees of continual learning.

Unfortunately \eqref{eq:muon-OGD} does not have an analytical form. To see this we can write out the Lagrangian of \eqref{eq:muon-OGD} and get
\begin{align*}
    &\textstyle \quad \cL(\Delta, \lambda) = \left\langle G + \sum_{i=1}^k\lambda_iC_i, \Delta\right\rangle\\
    \Longrightarrow &\quad \textstyle\Delta^\star(\lambda):=\argmin_{\|\Delta\|\le\eta} \cL(\Delta,\lambda) = -\eta ~\msgn\left(G + \sum_{i=1}^k\lambda_iC_i\right) \\
    \Longrightarrow &\quad \textstyle \cL(\Delta^\star(\lambda), \lambda) = -\eta \left\langle G + \sum_{i=1}^k\lambda_iC_i, \msgn\left(G + \sum_{i=1}^k\lambda_iC_i\right)\right\rangle = -\eta \underbrace{\textstyle\left\|G+\sum_{i=1}^k\lambda_iC_i\right\|_\star}_{:=f(\lambda)},
\end{align*}
where $\|\cdot\|_\star$ denote the nuclear norm. And thus we have that the dual problem for \eqref{eq:muon-OGD} is given by
\begin{equation}
\begin{split}\label{eq:muon-OGD-dual}
    \textstyle\min_\lambda f(\lambda) = \left\|G+\sum_{i=1}^k\lambda_iC_i\right\|_\star
\end{split}
\end{equation}
Unfortunately \eqref{eq:muon-OGD-dual} does not have a closed form solution, yet we can calculate the subgradient of the dual objective $f(\lambda)$ as:
 $\partial_{\lambda_i} f(\lambda) =  \left\langle C_i, \msgn\left(G + \sum_{i=1}^k\lambda_iC_i\right)\right\rangle,$
thus we can solve for the optimal $\lambda^\star$ iteratively via subgradient descent as
\begin{align*}
    \textstyle\lambda_i^{(m+1)} = \lambda_i^{(m)} - \eta_\lambda \left\langle C_i, \msgn\left(G + \sum_{i=1}^k\lambda_i^{(m)}C_i\right)\right\rangle
\end{align*}
and recover the solution of \eqref{eq:muon-OGD} by
 $\textstyle \Delta^\star = \msgn(G+\sum_{i=1}^k \lambda_i^\star C_i)$.
Thus we propose the following algorithm as in Algorithm \ref{alg:muon-ogd}. 
\begin{algorithm}[htbp]
\caption{Muon-OGD via Dual Iterations (Spectral-Norm Constrained Projection)}
\label{alg:muon-ogd}
\KwIn{Pretrained model parameters $\theta_0$; protected directions $\{C_i\}_{i=1}^k$ from the pretrained task; step size $\eta$; dual step size $\eta_\lambda$; inner iterations $T$; mini-batch sampler $\mathcal{B}(\cdot)$.}
\KwOut{Continually trained parameters $\theta$.}

\BlankLine
\textbf{Step 0 (Precompute constraints).} Obtain constraint matrices $\{C_i\}_{i=1}^k$ from the pretrained model and pretrained task (e.g., spanning a protected subspace or storing past-task gradient directions). Initialize $\theta_0, ~~\lambda_{-1} = \mathbf{0}\in\mathbb{R}^k,~~ G_{-1} = 0$.

\For{$t=0,\dots, T-1$}{
    Sample a mini-batch $\mathcal{S}_t\sim \mathcal{B}(\cdot)$ and compute momentum $G_t \leftarrow \beta G_{t-1} +\nabla_\theta \ell(\theta_t;\mathcal{S}_t) $
    
    \BlankLine
    \textbf{Inner loop (dual descent for \eqref{eq:muon-OGD-dual}).}
    Warmstart $\lambda^{(0)}$ from previous iteration $\lambda^{(0)} =\lambda_{t-1}$.\\
    \For{$m=0,\dots,T_{\mathrm{in}}-1$}{
        Form the shifted matrix $H^{(m)} = G + \sum_{i=1}^k \lambda_i^{(m)} C_i $.\\
        Compute approximate matrix sign via Newton-Schulz iteration
            $S^{(m)} \!=\! \NS\!\left(\!H^{(m)}\!\right)$.\\
            Dual update:
                $\lambda_i^{(m+1)} 
                \leftarrow 
                \lambda_i^{(m)} 
                - \eta_\lambda \,\langle C_i, S^{(m)}\rangle .$
    }
    Set $\lambda_t = \lambda^{(T_\mathrm{in})}$ and $H_t = G_t +\sum_{i=1}^k \lambda_{i,t} C_i$\\
    \BlankLine
    \textbf{Primal update (Muon-style step).}\quad 
        $\Delta_t \leftarrow -\eta~\NS(H_t), 
        \quad
        \theta_{t+1} \leftarrow \theta_t + \Delta_t .$
}
\end{algorithm}
\vspace{-5pt}

Algorithm~\ref{alg:muon-ogd} approximately solves the dual problem~\eqref{eq:muon-OGD-dual} via a small number of subgradient descent steps in the inner loop. The resulting dual variables induce a corrected gradient $H = G + \sum_i \lambda_i C_i$, which removes components that conflict with the constraint directions. The parameter update is then computed via a Muon-style step applied to $H$. In practice, the matrix sign $\msgn(\cdot)$ is efficiently approximated using a few Newton--Schulz iterations. Moreover, solving the dual problem to high accuracy is unnecessary: a small number of inner iterations (e.g., $T_{\mathrm{in}}\in\{1,2\}$) already yields an effective correction, resulting in an efficient single-loop algorithm.
\begin{remark}[Modularity]\label{rmk:modularity}
    The proposed framework is modular and can be viewed as a drop-in replacement for the update step in a wide class of projection-based continual learning methods. In particular, any method that specifies a collection of constraint directions $\{C_i\}$---whether derived from gradients, memory buffers, or structured subspaces---can be directly integrated into our formulation by replacing the standard Euclidean projection step with the spectral-norm-constrained update in~\eqref{eq:muon-OGD}. This suggests that existing projection-based approaches can be systematically upgraded to operate under spectral-norm geometry, potentially inheriting the conditioning and stability benefits observed in Muon, while preserving their original mechanisms for preventing catastrophic forgetting.
\end{remark}
\vspace{-5pt}
\subsection{Low-rank constraints from singular vectors}
\vspace{-5pt}
We illustrate a concrete instantiation of~\eqref{eq:muon-OGD} obtained by specifying the constraint directions $\{C_i\}$ using constructions from existing projection-based continual learning methods. In particular, recent works~\cite{nayak2025sculpting} extract constraint subspaces from the dominant singular components of a pretrained task. Adopting this approach, let $U\in\mathbb{R}^{m\times k}$ and $V\in\mathbb{R}^{n\times k}$ denote the top-$k$ left and right singular vectors obtained from the pretrained model or task statistics.  In this setting, the constraints in~\eqref{eq:muon-OGD} take the form
\vspace{-10pt}
\begin{align*}
\textstyle U^\top \Delta V = 0,
\end{align*}
which enforces orthogonality of $\Delta$ to the bilinear subspace spanned by $\{u_a v_b^\top\}_{a,b=1}^k$. Equivalently, this corresponds to choosing $C_i = C_{ab} = u_a v_b^\top$ in~\eqref{eq:muon-OGD}, so that the general constraint $\langle C_i,\Delta\rangle=0$ reduces to $\langle C_{ab},\Delta\rangle=0$ for all $(a,b)\in[k]\times[k]$. Under this parameterization, the dual variables can be organized into a matrix $\Lambda\in\mathbb{R}^{k\times k}$, yielding the structured correction
\begin{align*}
\textstyle \sum_{a,b} \Lambda_{ab} C_{ab} = U \Lambda V^\top.
\end{align*}
Substituting this structure into~\eqref{eq:muon-OGD} leads to a specialized instance of the Muon-OGD update, where the dual optimization is carried out over a $k^2$-dimensional space and all operations can be implemented using low-rank matrix products. The resulting algorithm (Algorithm~\ref{alg:muon-ogd-uv}) is thus a direct instantiation of the general framework with a structured choice of $\{C_i\}$, enabling more efficient computation while preserving the same constrained formulation.

\begin{algorithm}[htbp]
\caption{Muon-OGD with Low-Rank  Constraints}
\label{alg:muon-ogd-uv}
\KwIn{Pretrained model parameters $\theta_0$; rank $k$; step size $\eta$; dual step size $\eta_\Lambda$; inner iterations $T_{\mathrm{in}}$; mini-batch sampler $\mathcal{B}(\cdot)$.}
\KwOut{Continually trained parameters $\theta$.}

\BlankLine
\textbf{Step 0 (Precompute constraints).}
Obtain $U\in\mathbb{R}^{m\times k}$ and $V\in\mathbb{R}^{n\times k}$ as the top-$k$ left/right singular vectors from the pretrained task/model. 
Initialize $\theta_0$, $\Lambda_{-1} = \mathbf{0}\in\mathbb{R}^{k\times k}$, $G_{-1}=0$.

\For{$t=0,\dots,T-1$}{
    Sample a mini-batch $\mathcal{S}_t\sim \mathcal{B}(\cdot)$ and compute momentum 
    $G_t \leftarrow \beta G_{t-1} + \nabla_\theta \ell(\theta_t;\mathcal{S}_t)$.
    
    \BlankLine
    \textbf{Inner loop (dual descent).}\\
    Warm-start $\Lambda^{(0)} \leftarrow \Lambda_{t-1}$.\\
    \For{$m=0,\dots,T_{\mathrm{in}}-1$}{
        Form the shifted matrix 
        $H^{(m)} \leftarrow G_t + U\,\Lambda^{(m)} V^\top$.\\
        Compute approximate matrix sign via Newton--Schulz 
        $S^{(m)} \leftarrow \NS(H^{(m)})$.\\
        Dual update:
        $\Lambda^{(m+1)} \leftarrow \Lambda^{(m)} - \eta_\Lambda \, U^\top S^{(m)} V$.
    }

    Set $\Lambda_t \leftarrow \Lambda^{(T_{\mathrm{in}})}$ and 
    $H_t \leftarrow G_t + U \Lambda_t V^\top$.\\
    
    \BlankLine
    \textbf{Primal update (Muon-style step).}\quad 
    $\Delta_t \leftarrow -\eta\,\NS(H_t), \quad
    \theta_{t+1} \leftarrow \theta_t + \Delta_t$.
}
\end{algorithm}

In this paper, we focus on this specific instantiation for both practical and conceptual reasons. First, the resulting constraint subspace admits a structured low-rank representation, reducing the dimensionality of the dual problem and enabling efficient matrix--matrix implementations. Second, the formulation operates in the full-parameter fine-tuning regime, where updates are applied directly to matrix-valued weights and therefore align naturally with the matrix-centric geometry underlying Muon. In contrast, parameter-efficient fine-tuning (PEFT) methods such as LoRA restrict updates to low-rank adapter modules, extending it to these regimes would require reinterpreting the spectral-norm constraint under the induced low-rank parameterization. At the same time, as noted in Remark~\ref{rmk:modularity}, the proposed framework is modular and accommodates arbitrary choices of constraint directions $\{C_i\}$. Exploring alternative constructions of $\{C_i\}$ within the spectral-norm formulation, as well as extending the method to LoRA and other PEFT settings, remains an interesting direction for future work.

\vspace{-5pt}
\section{Experimental Results}
\label{sec:numerical_results}
\vspace{-5pt}
We evaluate Muon-OGD on both standard continual learning benchmarks and domain-specialized LLM adaptation settings. Our goal is to test whether the proposed spectral-norm-aware constrained update improves the stability--plasticity trade-off across different task lengths, domains, and model architectures. We compare against representative continual learning baselines, including recent orthogonal-gradient methods such as O-LoRA and Sculpting Subspaces. Unless otherwise specified, we follow standard task orders from prior work and report results averaged over multiple runs. Additional implementation details and hyperparameters are provided in Appendix~\ref{appendix_numerical_details}.

\vspace{-2pt}
\subsection{Benchmarks and Evaluation Protocol}
\label{subsec:benchmarks}
\vspace{-2pt}
We evaluate Muon-OGD on benchmarks with different task lengths, domain shifts, and model capabilities:

\begin{itemize}[leftmargin=15pt, itemsep=1pt, topsep=1pt]
    \item \textbf{Standard Continual Learning Benchmark (5 Tasks):} This benchmark contains five classic text classification tasks: AG News, Amazon Reviews, Yelp Reviews, DBpedia, and Yahoo Answers~\cite{zhang2015character}.
    
    \item \textbf{Extended Continual Learning Benchmark (15 Tasks):} This benchmark extends the standard setting to a longer task stream~\cite{razdaibiedina2023progressive}. It includes the original 5 tasks, four GLUE datasets (MNLI, QQP, RTE, SST-2)~\cite{wang2018glue}, five SuperGLUE datasets (WiC, CB, COPA, MultiRC, BoolQ)~\cite{wang2019superglue}, and IMDB.
    
    \item \textbf{TRACE Benchmark:} TRACE evaluates continual learning for instruction-tuned LLMs across diverse skills, including multilingual understanding, domain knowledge, arithmetic reasoning, and coding~\cite{wang2023trace}.
    
    \item \textbf{Domain-Specific Sequential Learning (Coding, Math, Medical):} To test specialized-domain adaptation, we train sequentially on BigCodeBench~\cite{zhuo2024bigcodebench}, GSM8K~\cite{cobbe2021training}, and HuatuoGPT-o1~\cite{chen2412huatuogpt}.
\end{itemize}

\paragraph{Models and Protocol.}
We evaluate both encoder-decoder and decoder-only architectures, including T5-Large, LLaMA2 (7B), Qwen2.5 (1.5B, 3B, 7B), LLaMA3.2 (1B, 3B). We follow standard continual learning protocols and average results over three independent runs with randomly permuted task sequences when applicable. Implementation details and hyperparameters are provided in Appendix~\ref{appendix_numerical_details} .

\vspace{-5pt}
\subsection{Baseline Methods}
\label{subsec:baselines}
\vspace{-5pt}
We compare Muon-OGD with representative baselines from several continual learning paradigms.

\begin{itemize}[leftmargin=15pt, itemsep=1pt, topsep=1pt]
    \item \textbf{Lower Bound (SeqFT):} Sequential full-model fine-tuning without replay or anti-forgetting constraints, representing standard catastrophic forgetting behavior.
    
    \item \textbf{Parameter-Efficient Fine-Tuning (PEFT):} LoRA-based continual learning methods, including \textbf{SeqLoRA}, \textbf{IncLoRA}, and \textbf{O-LoRA}.
    
    \item \textbf{Replay-Based Methods:} Experience \textbf{Replay}, which trains on the current task together with stored samples from previous tasks.
    
    \item \textbf{Regularization Methods:} \textbf{Elastic Weight Consolidation (EWC)}~\cite{kirkpatrick2017overcoming}, which penalizes changes to parameters important for previous tasks.
    
    \item \textbf{Prompt-Based Methods:} Dynamic prompting approaches, including \textbf{L2P}~\cite{wang2022learning} and \textbf{ProgPrompt}~\cite{razdaibiedina2023progressive}.
    
    \item \textbf{Upper Bounds (PerTaskFT \& MTL):} \textbf{PerTaskFT} trains a separate model per task, \textbf{MTL} trains jointly on all tasks. These serve as performance ceilings outside the continual learning setting.
\end{itemize}

\vspace{-5pt}
\subsection{Main Result}
\vspace{-5pt}
\begin{table*}[h]
\centering
\scriptsize
\setlength{\tabcolsep}{4.0pt}
\renewcommand{\arraystretch}{1.02}
\caption{
Comparison of Average Accuracy (\%) on standard and 15-task continual learning benchmarks using T5-Large.
Our Muon-OGD results are averaged over three seeds for each order.
For each column among continual learning methods, the best performance is highlighted in \textbf{bold}, and the second-best is \underline{underlined}.
}
\label{tab:t5_large_cl_results}
\resizebox{0.88\textwidth}{!}{
\begin{tabular}{@{}lcccccccc@{}}
\toprule
\multirow{2}{*}{\textbf{Method}}
& \multicolumn{4}{c}{\textbf{Standard CL Benchmark}}
& \multicolumn{4}{c}{\textbf{15-Task CL Benchmark}} \\
\cmidrule(lr){2-5}
\cmidrule(l){6-9}
& \textbf{Order-1} & \textbf{Order-2} & \textbf{Order-3} & \textbf{Avg}
& \textbf{Order-4} & \textbf{Order-5} & \textbf{Order-6} & \textbf{Avg} \\
\midrule
SeqFT      & 18.9 & 24.9 & 41.7 & 28.5 &  7.4 &  7.4 &  7.5 &  7.4 \\
SeqLoRA    & 44.6 & 32.7 & 53.7 & 43.7 &  2.3 &  0.6 &  1.9 &  1.6 \\
IncLoRA    & 66.0 & 64.9 & 68.3 & 66.4 & 63.3 & 58.5 & 61.7 & 61.2 \\
Replay     & 55.2 & 56.9 & 61.3 & 57.8 & 55.0 & 54.6 & 53.1 & 54.2 \\
EWC        & 48.7 & 47.7 & 54.5 & 50.3 & 45.3 & 44.5 & 45.6 & 45.1 \\
LwF        & 54.4 & 53.1 & 49.6 & 52.3 & 50.1 & 43.1 & 47.4 & 46.9 \\
L2P        & 60.3 & 61.7 & 61.1 & 60.7 & 57.5 & 53.8 & 56.9 & 56.1 \\
LFPT5      & 67.6 & 72.6 & 77.9 & 72.7 & 70.4 & 68.2 & 69.1 & 69.2 \\
O-LoRA     & \underline{75.4} & 75.7 & 76.3 & 75.8 & \textbf{72.3} & 64.8 & 71.6 & 69.6 \\
OSFT       & 75.3 & \underline{74.0} & \textbf{78.4} & \underline{75.9}
           & 71.6 & \underline{69.6} & \underline{72.7} & \underline{71.3} \\
\textbf{Muon-OGD (Ours)}
           & \textbf{78.9$_{\pm0.8}$} & \textbf{79.4$_{\pm0.7}$} & \underline{78.3$_{\pm1.0}$} & \textbf{78.9$_{\pm1.0}$}
           & \underline{72.2$_{\pm0.8}$} & \textbf{{70.3$_{\pm0.7}$}} & \textbf{73.5$_{\pm0.4}$} & \textbf{{72.0$_{\pm1.5}$}} \\
\midrule
SLERP      & 40.5 & 43.0 & 45.8 & 43.1 &  2.4 &  1.5 &  2.7 &  2.2 \\
TIES       & 35.0 & 38.5 & 37.8 & 37.1 &  7.8 &  7.1 &  5.8 &  6.9 \\
\midrule
ProgPrompt & 75.2 & 75.0 & 75.1 & 75.1 & 78.0 & 77.7 & 77.9 & 77.9 \\
PerTaskFT  & 70.0 & 70.0 & 70.0 & 70.0 & 78.1 & 78.1 & 78.1 & 78.1 \\
MTL (Upper Bound)
           & 80.0 & 80.0 & 80.0 & 80.0 & 76.5 & 76.5 & 76.5 & 76.5 \\
\bottomrule
\end{tabular}
}
\end{table*}

Table~\ref{tab:t5_large_cl_results} compares Muon-OGD with representative continual learning baselines on the T5-Large continual learning benchmarks. 
We consider both the standard benchmark and the large 15-task benchmark, following the same experimental protocol as O-LoRA. 
On the standard benchmark, Muon-OGD achieves the best average accuracy among the compared continual learning methods, reaching $78.9\%$ and improving over O-LoRA ($75.8\%$) and OSFT ($75.9\%$). 
On the large benchmark, Muon-OGD remains highly competitive under a longer and more heterogeneous task stream, achieving $72.0\%$ average accuracy and improving over O-LoRA ($69.6\%$) and OSFT ($71.3\%$). 
Across the six T5 task orders, Muon-OGD obtains the best result in most orders and remains close to the strongest baseline in the remaining cases, suggesting that the proposed operator-norm-aware update is effective across both short and long continual learning streams.


\begin{table}[htbp]
\centering
\small
\setlength{\tabcolsep}{7pt}
\renewcommand{\arraystretch}{1.12}
\caption{
TRACE benchmark performance with the LLaMA-2-7B-Chat backbone model.
}
\label{tab:trace_backbones}
\begin{tabular}{@{}lcc@{}}
\toprule
\multirow{2}{*}{\textbf{Method}}
& \multicolumn{2}{c}{\textbf{LLaMA-2-7B-Chat}} \\
\cmidrule(l){2-3} 
& \textbf{AA (\%)} & \textbf{BT (\%)} \\
\midrule
SeqFT        & 23.0 &  -8.3 \\
LoRASeqFT    &  9.2 & -24.6 \\
O-LoRA       & 41.3 &  -6.2 \\
OSFT         & 48.4 &  -7.1 \\
\textbf{Ours}
             & \textbf{49.4$_{\pm0.3}$} & \textbf{-4.7$_{\pm1.2}$} \\
\midrule
PerTaskFT    & 57.6 & N/A  \\
MTL          & 52.3 & N/A  \\
\bottomrule
\end{tabular}

\footnotesize

\end{table}

Table~\ref{tab:trace_backbones} evaluates Muon-OGD on the TRACE benchmark using the decoder-only LLaMA-2-7B-Chat backbone. 
TRACE contains a diverse sequence of instruction-following tasks, making it substantially different from the T5-Large classification-style benchmarks. 
Muon-OGD achieves the best performance among all compared methods, with $49.4\%$ AA and $-4.7\%$ BT representing the highest final average accuracy and lowest forgetting rate.

\begin{table}[t]
\centering
\scriptsize
\setlength{\tabcolsep}{2.6pt}
\renewcommand{\arraystretch}{1.02}
\caption{
Continual learning performance of Qwen2.5-1.5B-Instruct and Qwen2.5-3B-Instruct
under the sequential curriculum Coding $\rightarrow$ Math $\rightarrow$ Medical.
All values are testing accuracy percentages (\%). For each metric within each stage,
the best result is highlighted in \textbf{bold} and the second-best is \underline{underlined}.
AA denotes the average accuracy over observed tasks.
}
\label{tab:combined_sequence_results}
\resizebox{\textwidth}{!}{
\begin{tabular}{@{}llcccccccc@{}}
\toprule
\multirow{2}{*}{\textbf{Stage}} 
& \multirow{2}{*}{\textbf{Method}}
& \multicolumn{4}{c}{\textbf{Qwen2.5-1.5B-Instruct}}
& \multicolumn{4}{c}{\textbf{Qwen2.5-3B-Instruct}} \\
\cmidrule(lr){3-6} \cmidrule(l){7-10}
& 
& \textbf{Code} & \textbf{Math} & \textbf{Med.} & \textbf{AA}
& \textbf{Code} & \textbf{Math} & \textbf{Med.} & \textbf{AA} \\
\midrule

\multirow{4}{*}{\textbf{A}}
& SeqSFT 
& 17.1 & -- & -- & 17.1
& 31.1 & -- & -- & 31.1 \\
& O-LoRA 
& \underline{17.9} & -- & -- & \underline{17.9}
& \underline{32.7} & -- & -- & \underline{32.7} \\
& Sculpt. 
& 17.1 & -- & -- & 17.1
& \underline{32.7} & -- & -- & \underline{32.7} \\
& \textbf{Muon-OGD}
& \textbf{18.4}$_{\pm0.2}$ & -- & -- & \textbf{18.4}$_{\pm0.2}$
& \textbf{32.9}$_{\pm0.4}$ & -- & -- & \textbf{32.9}$_{\pm0.4}$ \\

\midrule

\multirow{4}{*}{\textbf{B}}
& SeqSFT 
& 14.1 & 46.5 & -- & 30.3
& 18.2 & 60.6 & -- & 39.4 \\
& O-LoRA 
& \underline{17.6} & \underline{51.8} & -- & \underline{34.7}
& 18.7 & \textbf{67.0} & -- & \underline{42.9} \\
& Sculpt. 
& 13.8 & 47.0 & -- & 30.4
& \underline{20.8} & 62.6 & -- & 41.7 \\
& \textbf{Muon-OGD}
& \textbf{17.8}$_{\pm0.6}$ & \textbf{52.5}$_{\pm0.7}$ & -- & \textbf{35.2}$_{\pm0.5}$
& \textbf{29.7}$_{\pm0.4}$ & \underline{64.6}$_{\pm0.2}$ & -- & \textbf{47.2}$_{\pm0.2}$ \\

\midrule

\multirow{4}{*}{\textbf{C}}
& SeqSFT 
& \textbf{15.0} & 46.4 & 22.0 & 27.8
& 19.1 & 62.8 & \underline{28.0} & 36.6 \\
& O-LoRA 
& 14.8 & 46.4 & 20.8 & 27.3
& 16.2 & \underline{63.0} & 25.2 & 34.8 \\
& Sculpt. 
& 14.8 & \underline{47.4} & \underline{22.6} & \underline{28.3}
& \underline{32.1} & \underline{63.0} & 27.6 & \underline{40.9} \\
& \textbf{Muon-OGD}
& \underline{14.9}$_{\pm0.2}$ & \textbf{51.0}$_{\pm0.3}$ & \textbf{23.4}$_{\pm0.8}$ & \textbf{29.8}$_{\pm0.3}$
& \textbf{32.9}$_{\pm0.4}$ & \textbf{69.8}$_{\pm0.7}$ & \textbf{28.8}$_{\pm0.8}$ & \textbf{43.8}$_{\pm0.4}$ \\

\bottomrule
\vspace{-5pt}
\end{tabular}
}
\end{table}
Table~\ref{tab:combined_sequence_results} further evaluates Muon-OGD on a domain-specific curriculum, Coding $\rightarrow$ Math $\rightarrow$ Medical, using Qwen2.5-1.5B-Instruct and Qwen2.5-3B-Instruct. 
This benchmark tests whether the method can preserve earlier domain skills while adapting to substantially different reasoning and output formats. 
Muon-OGD achieves the highest average accuracy at every stage for both model sizes. 
In particular, after the final Medical stage, Muon-OGD reaches $29.8\%$ AA on Qwen2.5-1.5B-Instruct and $43.8\%$ AA on Qwen2.5-3B-Instruct, outperforming Sculpting Subspaces and O-LoRA. 
These results show that Muon-OGD remains effective in heterogeneous domain-adaptation scenarios where tasks differ substantially in knowledge domain, reasoning style, and output format.

Overall, the main results support three conclusions. 
First, replacing Euclidean-style projection with operator-norm-aware Muon updates improves continual learning across standard T5-Large CL streams, large long-sequence CL streams, TRACE instruction-following tasks, and domain-specific curricula. 
Second, Muon-OGD reduces forgetting while maintaining strong plasticity, as reflected by its improved AA and BT on TRACE. 
Third, the method is compatible with both encoder--decoder and decoder-only architectures, suggesting that the proposed update geometry is broadly applicable across different LLM backbones and continual learning settings.

\vspace{-5pt}
\subsection{Ablation Study}
\label{subsec:main_ablation}
\vspace{-5pt}
We summarize the main ablation findings here and provide full details in Appendix~\ref{sec:ablation}. Overall, Muon-OGD is most effective when plasticity and stability are properly balanced. The Muon step size and AdamW learning rate need to be large enough to support adaptation to the current task, while the dual correction and protected subspace rank need to be strong enough to limit interference with previous tasks. In our experiments, overly small learning rates reduce current-task learning, whereas overly aggressive dual updates can over-constrain the update direction and hurt plasticity. 
\vspace{-5pt}
\section{Conclusion and Future Work}
\label{sec:conclusion}
\vspace{-5pt}
In this paper, we introduced a novel continual learning framework combining the operator-norm--aware geometry of the Muon optimizer with projection-based stability updates. By solving a spectral-norm–aware constrained optimization problem, we effectively preserve performance on previous domains without sacrificing the computational benefits of momentum orthogonalized by Newton-Schulz iterations. Our future work will focus on efficiently identifying the optimal constraint directions $C_i$ by exploring the most dominant singular values of the parameter subspaces \cite{wang2023orthogonalsubspacelearninglanguage,qiao2024learn}. We will also focus on scaling the algorithmic implementation for larger pre-trained language models and analyzing its theoretical convergence guarantees under varied sequential learning settings.


\bibliographystyle{unsrtnat}  
\bibliography{references}

@article{parisi2019continual,
  title={Continual lifelong learning with neural networks: A review},
  author={Parisi, German I and Kemker, Ronald and Part, Jose L and Kanan, Christopher and Wermter, Stefan},
  journal={Neural networks},
  volume={113},
  pages={54--71},
  year={2019},
  publisher={Elsevier}
}

@article{de2021continual,
  title={A continual learning survey: Defying forgetting in classification tasks},
  author={De Lange, Matthias and Aljundi, Rahaf and Masana, Marc and Parisot, Sarah and Jia, Xu and Leonardis, Ale{\v{s}} and Slabaugh, Gregory and Tuytelaars, Tinne},
  journal={IEEE transactions on pattern analysis and machine intelligence},
  volume={44},
  number={7},
  pages={3366--3385},
  year={2021},
  publisher={IEEE}
}

@article{wang2024comprehensive,
  title={A comprehensive survey of continual learning: Theory, method and application},
  author={Wang, Liyuan and Zhang, Xingxing and Su, Hang and Zhu, Jun},
  journal={IEEE transactions on pattern analysis and machine intelligence},
  volume={46},
  number={8},
  pages={5362--5383},
  year={2024},
  publisher={IEEE}
}

@article{kirkpatrick2017overcoming,
  title={Overcoming catastrophic forgetting in neural networks},
  author={Kirkpatrick, James and Pascanu, Razvan and Rabinowitz, Neil and Veness, Joel and Desjardins, Guillaume and Rusu, Andrei A and Milan, Kieran and Quan, John and Ramalho, Tiago and Grabska-Barwinska, Agnieszka and others},
  journal={Proceedings of the national academy of sciences},
  volume={114},
  number={13},
  pages={3521--3526},
  year={2017},
  publisher={National Academy of Sciences}
}

@article{li2017learning,
  title={Learning without forgetting},
  author={Li, Zhizhong and Hoiem, Derek},
  journal={IEEE transactions on pattern analysis and machine intelligence},
  volume={40},
  number={12},
  pages={2935--2947},
  year={2017},
  publisher={IEEE}
}

@inproceedings{zenke2017continual,
  title={Continual learning through synaptic intelligence},
  author={Zenke, Friedemann and Poole, Ben and Ganguli, Surya},
  booktitle={International conference on machine learning},
  pages={3987--3995},
  year={2017},
  organization={Pmlr}
}

@misc{bernstein2025deriving,
  author = {Jeremy Bernstein},
  title = {Deriving Muon},
  url = {https://jeremybernste.in/writing/deriving-muon},
  year = {2025}
}

@inproceedings{aljundi2018memory,
  title={Memory aware synapses: Learning what (not) to forget},
  author={Aljundi, Rahaf and Babiloni, Francesca and Elhoseiny, Mohamed and Rohrbach, Marcus and Tuytelaars, Tinne},
  booktitle={Proceedings of the European conference on computer vision (ECCV)},
  pages={139--154},
  year={2018}
}

@article{lopez2017gradient,
  title={Gradient episodic memory for continual learning},
  author={Lopez-Paz, David and Ranzato, Marc'Aurelio},
  journal={Advances in neural information processing systems},
  volume={30},
  year={2017}
}

@article{chaudhry2018efficient,
  title={Efficient lifelong learning with a-gem},
  author={Chaudhry, Arslan and Ranzato, Marc'Aurelio and Rohrbach, Marcus and Elhoseiny, Mohamed},
  journal={arXiv preprint arXiv:1812.00420},
  year={2018}
}

@inproceedings{farajtabar2020orthogonal,
  title={Orthogonal gradient descent for continual learning},
  author={Farajtabar, Mehrdad and Azizan, Navid and Mott, Alex and Li, Ang},
  booktitle={International conference on artificial intelligence and statistics},
  pages={3762--3773},
  year={2020},
  organization={PMLR}
}

@article{saha2021gradient,
  title={Gradient projection memory for continual learning},
  author={Saha, Gobinda and Garg, Isha and Roy, Kaushik},
  journal={arXiv preprint arXiv:2103.09762},
  year={2021}
}

@inproceedings{wang2022learning,
  title={Learning to prompt for continual learning},
  author={Wang, Zifeng and Zhang, Zizhao and Lee, Chen-Yu and Zhang, Han and Sun, Ruoxi and Ren, Xiaoqi and Su, Guolong and Perot, Vincent and Dy, Jennifer and Pfister, Tomas},
  booktitle={Proceedings of the IEEE/CVF conference on computer vision and pattern recognition},
  pages={139--149},
  year={2022}
}

@inproceedings{wang2022dualprompt,
  title={Dualprompt: Complementary prompting for rehearsal-free continual learning},
  author={Wang, Zifeng and Zhang, Zizhao and Ebrahimi, Sayna and Sun, Ruoxi and Zhang, Han and Lee, Chen-Yu and Ren, Xiaoqi and Su, Guolong and Perot, Vincent and Dy, Jennifer and others},
  booktitle={European conference on computer vision},
  pages={631--648},
  year={2022},
  organization={Springer}
}

@article{razdaibiedina2023progressive,
  title={Progressive prompts: Continual learning for language models},
  author={Razdaibiedina, Anastasia and Mao, Yuning and Hou, Rui and Khabsa, Madian and Lewis, Mike and Almahairi, Amjad},
  journal={arXiv preprint arXiv:2301.12314},
  year={2023}
}

@article{qin2021lfpt5,
  title={Lfpt5: A unified framework for lifelong few-shot language learning based on prompt tuning of t5},
  author={Qin, Chengwei and Joty, Shafiq},
  journal={arXiv preprint arXiv:2110.07298},
  year={2021}
}

@article{hu2022lora,
  title={Lora: Low-rank adaptation of large language models.},
  author={Hu, Edward J and Shen, Yelong and Wallis, Phillip and Allen-Zhu, Zeyuan and Li, Yuanzhi and Wang, Shean and Wang, Liang and Chen, Weizhu and others},
  journal={Iclr},
  volume={1},
  number={2},
  pages={3},
  year={2022}
}

@article{bernstein2025modular,
  title={Modular duality in deep learning},
  author={Bernstein, Jeremy and Newhouse, Laker},
  journal={arXiv preprint arXiv:2410.21265},
  year={2024}
}

@article{chen2025muon,
  title={Muon optimizes under spectral norm constraints},
  author={Chen, Lizhang and Li, Jonathan and Liu, Qiang},
  journal={arXiv preprint arXiv:2506.15054},
  year={2025}
}

@inproceedings{wang2023orthogonal,
  title={Orthogonal subspace learning for language model continual learning},
  author={Wang, Xiao and Chen, Tianze and Ge, Qiming and Xia, Han and Bao, Rong and Zheng, Rui and Zhang, Qi and Gui, Tao and Huang, Xuan-Jing},
  booktitle={Findings of the Association for Computational Linguistics: EMNLP 2023},
  pages={10658--10671},
  year={2023}
}

@article{nayak2025sculpting,
  title={Sculpting subspaces: Constrained full fine-tuning in llms for continual learning},
  author={Nayak, Nikhil Shivakumar and Killamsetty, Krishnateja and Han, Ligong and Bhandwaldar, Abhishek and Chanda, Prateek and Xu, Kai and Wang, Hao and Pareja, Aldo and Silkin, Oleg and Eyceoz, Mustafa and others},
  journal={arXiv preprint arXiv:2504.07097},
  year={2025}
}

@article{su2025isotropic,
  title={Isotropic Curvature Model for Understanding Deep Learning Optimization: Is Gradient Orthogonalization Optimal?},
  author={Su, Weijie},
  journal={arXiv preprint arXiv:2511.00674},
  year={2025}
}

@inproceedings{xiong2026oplora,
  title={Oplora: Orthogonal projection lora prevents catastrophic forgetting during parameter-efficient fine-tuning},
  author={Xiong, Yifeng and Xie, Xiaohui},
  booktitle={Proceedings of the AAAI Conference on Artificial Intelligence},
  volume={40},
  number={40},
  pages={34088--34096},
  year={2026}
}

@misc{zhang2025dynamicorthogonalcontinualfinetuning,
      title={Dynamic Orthogonal Continual Fine-tuning for Mitigating Catastrophic Forgettings}, 
      author={Zhixin Zhang and Zeming Wei and Meng Sun},
      year={2025},
      eprint={2509.23893},
      archivePrefix={arXiv},
      primaryClass={cs.LG},
      url={https://arxiv.org/abs/2509.23893}, 
}

@InProceedings{farajtabar20orthogonal,
  title = 	 {Orthogonal Gradient Descent for Continual Learning},
  author =       {Farajtabar, Mehrdad and Azizan, Navid and Mott, Alex and Li, Ang},
  booktitle = 	 {Proceedings of the Twenty Third International Conference on Artificial Intelligence and Statistics},
  pages = 	 {3762--3773},
  year = 	 {2020},
  editor = 	 {Chiappa, Silvia and Calandra, Roberto},
  volume = 	 {108},
  series = 	 {Proceedings of Machine Learning Research},
  month = 	 {26--28 Aug},
  publisher =    {PMLR},
  url = 	 {https://proceedings.mlr.press/v108/farajtabar20a.html}
}

@article{liu2025muon,
  title={Muon is scalable for LLM training},
  author={Liu, Jingyuan and Su, Jianlin and Yao, Xingcheng and Jiang, Zhejun and Lai, Guokun and Du, Yulun and Qin, Yidao and Xu, Weixin and Lu, Enzhe and Yan, Junjie and others},
  journal={arXiv preprint arXiv:2502.16982},
  year={2025}
}

@online{jordan_muon,
  author = {Jordan, Keller},
  title = {Muon: An optimizer for hidden layers in neural networks
},
  url = {https://kellerjordan.github.io/posts/muon/},
  note = {Blog post},
}

@misc{wang2023orthogonalsubspacelearninglanguage,
      title={Orthogonal Subspace Learning for Language Model Continual Learning}, 
      author={Xiao Wang and Tianze Chen and Qiming Ge and Han Xia and Rong Bao and Rui Zheng and Qi Zhang and Tao Gui and Xuanjing Huang},
      year={2023},
      eprint={2310.14152},
      archivePrefix={arXiv},
      primaryClass={cs.CL},
      url={https://arxiv.org/abs/2310.14152}, 
}

@inproceedings{
qiao2024learn,
title={Learn more, but bother less: parameter efficient continual learning},
author={Fuli Qiao and Mehrdad Mahdavi},
booktitle={The Thirty-eighth Annual Conference on Neural Information Processing Systems},
year={2024},
url={https://openreview.net/forum?id=ZxtaNh5UYB}
}

@article{zhang2015character,
  title={Character-level convolutional networks for text classification},
  author={Zhang, Xiang and Zhao, Junbo and LeCun, Yann},
  journal={Advances in neural information processing systems},
  volume={28},
  year={2015}
}

@inproceedings{wang2018glue,
  title={GLUE: A multi-task benchmark and analysis platform for natural language understanding},
  author={Wang, Alex and Singh, Amanpreet and Michael, Julian and Hill, Felix and Levy, Omer and Bowman, Samuel},
  booktitle={Proceedings of the 2018 EMNLP workshop BlackboxNLP: Analyzing and interpreting neural networks for NLP},
  pages={353--355},
  year={2018}
}

@article{wang2019superglue,
  title={Superglue: A stickier benchmark for general-purpose language understanding systems},
  author={Wang, Alex and Pruksachatkun, Yada and Nangia, Nikita and Singh, Amanpreet and Michael, Julian and Hill, Felix and Levy, Omer and Bowman, Samuel},
  journal={Advances in neural information processing systems},
  volume={32},
  year={2019}
}

@article{wang2023trace,
  title={Trace: A comprehensive benchmark for continual learning in large language models},
  author={Wang, Xiao and Zhang, Yuansen and Chen, Tianze and Gao, Songyang and Jin, Senjie and Yang, Xianjun and Xi, Zhiheng and Zheng, Rui and Zou, Yicheng and Gui, Tao and others},
  journal={arXiv preprint arXiv:2310.06762},
  year={2023}
}

@article{zhuo2024bigcodebench,
  title={Bigcodebench: Benchmarking code generation with diverse function calls and complex instructions},
  author={Zhuo, Terry Yue and Vu, Minh Chien and Chim, Jenny and Hu, Han and Yu, Wenhao and Widyasari, Ratnadira and Yusuf, Imam Nur Bani and Zhan, Haolan and He, Junda and Paul, Indraneil and others},
  journal={arXiv preprint arXiv:2406.15877},
  year={2024}
}

@article{cobbe2021training,
  title={Training verifiers to solve math word problems},
  author={Cobbe, Karl and Kosaraju, Vineet and Bavarian, Mohammad and Chen, Mark and Jun, Heewoo and Kaiser, Lukasz and Plappert, Matthias and Tworek, Jerry and Hilton, Jacob and Nakano, Reiichiro and others},
  journal={arXiv preprint arXiv:2110.14168},
  year={2021}
}

@article{chen2412huatuogpt,
  title={Huatuogpt-o1, towards medical complex reasoning with llms, 2024},
  author={Chen, Junying and Cai, Zhenyang and Ji, Ke and Wang, Xidong and Liu, Wanlong and Wang, Rongsheng and Hou, Jianye and Wang, Benyou},
  journal={URL https://arxiv. org/abs/2412.18925}
}

@article{shin2017continual,
  title={Continual learning with deep generative replay},
  author={Shin, Hanul and Lee, Jung Kwon and Kim, Jaehong and Kim, Jiwon},
  journal={Advances in neural information processing systems},
  volume={30},
  year={2017}
}

@inproceedings{yoon2018lifelong,
  title={Lifelong learning with dynamically expandable networks},
  author={Yoon, Jaehong and Yang, Eunho and Lee, Jeongtae and Hwang, Sung Ju},
  booktitle={6th International Conference on Learning Representations, ICLR 2018},
  year={2018}
}

@article{goodfellow2013empirical,
  title={An empirical investigation of catastrophic forgetting in gradient-based neural networks},
  author={Goodfellow, Ian J and Mirza, Mehdi and Xiao, Da and Courville, Aaron and Bengio, Yoshua},
  journal={arXiv preprint arXiv:1312.6211},
  year={2013}
}

@article{shen2025convergence,
  title={On the convergence analysis of muon},
  author={Shen, Wei and Huang, Ruichuan and Huang, Minhui and Shen, Cong and Zhang, Jiawei},
  journal={arXiv preprint arXiv:2505.23737},
  year={2025}
}

@article{loshchilov2017decoupled,
  title={Decoupled weight decay regularization},
  author={Loshchilov, Ilya and Hutter, Frank},
  journal={arXiv preprint arXiv:1711.05101},
  year={2017}
}

@article{davis2025spectral,
  title={When do spectral gradient updates help in deep learning?},
  author={Davis, Damek and Drusvyatskiy, Dmitriy},
  journal={arXiv preprint arXiv:2512.04299},
  year={2025}
}

@inproceedings{tekmen2025gem,
  title={GEM-Style Constraints for PEFT with Dual Gradient Projection in LoRA},
  author={Tekmen, Brian and Yin, Jason and Tong, Qianqian},
  booktitle={2025 IEEE International Conference on Data Mining Workshops (ICDMW)},
  pages={2736--2743},
  year={2025},
  organization={IEEE}
}

@inproceedings{bernsteinmodular,
  title={Modular Duality in Deep Learning},
  author={Bernstein, Jeremy and Newhouse, Laker},
  booktitle={Forty-second International Conference on Machine Learning}
}

@article{higham1995matrix,
  title={Matrix procrustes problems},
  author={Higham, Nick and Papadimitriou, Pythagoras},
  journal={Rapport technique, University of Manchester},
  year={1995}
}


\appendix


\section*{Declaration of LLM Usage.}
We used large language models as writing and coding assistants during the preparation of this manuscript. Specifically, LLMs were used to improve language clarity and presentation, assist with coding and debugging for numerical experiments. All technical ideas, theoretical results, experimental design, and final interpretations were developed, verified, and approved by the authors.

\section{Related Work (Detailed)}
\label{app:related_work}

\paragraph{Continual Learning in Large Language Models.}
Continual learning (CL) aims to enable models to acquire a sequence of new tasks while preserving performance on previously learned ones~\cite{parisi2019continual,de2021continual,wang2024comprehensive}. Classical CL methods are commonly grouped into regularization-based, replay-based, and architecture- or prompt-based approaches. Regularization-based methods, such as EWC~\cite{kirkpatrick2017overcoming}, Synaptic Intelligence~\cite{zenke2017continual}, MAS~\cite{aljundi2018memory}, and Learning without Forgetting~\cite{li2017learning}, discourage updates that strongly perturb parameters or predictions important to earlier tasks. Replay-based methods, such as GEM and A-GEM~\cite{lopez2017gradient,chaudhry2018efficient}, mitigate forgetting by storing or replaying examples from previous tasks, but this can introduce memory, privacy, or data-governance concerns when applied to large-scale language models. Prompt- and parameter-efficient methods, including L2P~\cite{wang2022learning}, DualPrompt~\cite{wang2022dualprompt}, Progressive Prompts~\cite{razdaibiedina2023progressive}, LFPT5~\cite{qin2021lfpt5}, and LoRA-based adaptation~\cite{hu2022lora}, reduce the number of trainable parameters by adapting small prompt vectors or low-rank modules while keeping most pretrained weights fixed.

Recent work has shown that CL for LLMs is especially challenging because sequential instruction tuning can degrade not only previous task accuracy but also general reasoning, instruction-following, and safety behavior. TRACE~\cite{wang2023trace} provides a comprehensive benchmark for LLM continual learning, covering domain-specific tasks, multilingual ability, code generation, and mathematical reasoning in a unified evaluation format. Other language-model CL studies evaluate both standard continual-learning task sequences and longer multi-task sequences to measure the stability-plasticity trade-off under more realistic task streams~\cite{razdaibiedina2023progressive,wang2023orthogonal,nayak2025sculpting}. These benchmarks motivate update mechanisms that can learn new task-specific information while explicitly controlling interference with previously acquired capabilities.

\paragraph{Projection-based Continual Learning.}
Projection-based CL methods address catastrophic forgetting by modifying the update direction rather than only adding scalar penalties to the loss. GEM~\cite{lopez2017gradient} and A-GEM~\cite{chaudhry2018efficient} project or approximate-project gradients so that the loss on stored previous-task samples does not increase. Orthogonal Gradient Descent (OGD)~\cite{farajtabar2020orthogonal} instead projects the gradient for a new task onto a subspace that is approximately orthogonal to directions affecting previous-task outputs. Related orthogonal-subspace methods further enforce non-interference by constructing task-specific subspaces or decomposing representation spaces~\cite{saha2021gradient,wang2023orthogonal}. Compared with replay-based approaches, these methods reduce or avoid dependence on storing previous training data, but maintaining and projecting against historical gradient or representation subspaces can become expensive for modern LLMs.

For language-model continual learning, O-LoRA adapts the projection principle to low-rank adaptation~\cite{wang2023orthogonal}. Instead of updating the full model, O-LoRA assigns different tasks to low-rank subspaces and encourages these subspaces to remain mutually orthogonal. This design reduces task interference with only marginal additional parameters and avoids storing previous task data. However, because O-LoRA operates within low-rank adapters, its expressivity may be limited when full-model adaptation is needed. More recently, Sculpting Subspaces and Orthogonal Subspace Fine-Tuning (OSFT) propose constrained full fine-tuning strategies for LLM continual learning~\cite{nayak2025sculpting}. These methods use singular-value-based subspace analysis to identify task-relevant directions and constrain future updates away from critical directions associated with prior tasks. This line of work shows that geometric constraints and orthogonal update rules are effective for mitigating forgetting, but existing methods often rely on Euclidean projection or SVD-based subspace construction. In contrast, our method combines orthogonal projection with a Muon-style operator-norm-aware update geometry, aiming to preserve prior-task subspaces while improving the conditioning and stability of full-parameter updates.

\paragraph{Spectral-Norm-Aware Optimization and Muon.}
Standard LLM training and fine-tuning commonly use AdamW~\cite{loshchilov2017decoupled}, which applies coordinate-wise adaptive scaling with decoupled weight decay. Muon, short for MomentUm Orthogonalized by Newton--Schulz, is a recent optimizer designed for matrix-valued neural-network parameters~\cite{jordan_muon}. Unlike AdamW, Muon first forms a momentum update and then approximately orthogonalizes the matrix update using Newton--Schulz iterations, producing update directions close to the polar factor of the momentum matrix. Recent empirical work shows that Muon can scale to large language model training and may improve training efficiency compared with AdamW in large-scale pretraining regimes~\cite{liu2025muon}. From a theoretical perspective, modular duality and related analyses interpret Muon-style updates through operator-norm-aware geometry, while recent work further formalizes Muon as implicitly optimizing under spectral-norm constraints~\cite{bernstein2025modular,chen2025muon}. These views support the idea that matrix-valued updates should not always be treated as flat Euclidean vectors.

Beyond pretraining, Muon-style orthogonalized updates have also motivated work on parameter-efficient adaptation. For example, recent LoRA-oriented extensions formulate low-rank adaptation from a geometric or Riemannian perspective and use Muon-inspired orthogonalization to reduce parametrization imbalance in low-rank factors. These developments suggest that orthogonalized matrix updates are useful not only for optimization speed but also for controlling the geometry of learned parameter changes. However, existing Muon work primarily focuses on pretraining efficiency, optimization theory, or low-rank adaptation, rather than continual learning. Our work connects Muon-style orthogonalized updates with projection-based continual learning: by applying operator-norm-aware updates under orthogonal subspace constraints, we aim to reduce catastrophic forgetting while retaining the plasticity needed for sequential LLM fine-tuning.

\section{Limitations and Future Work}
\label{app:limitations}

Our study focuses on validating Muon-OGD in representative continual learning settings for language models, and several directions remain open for further exploration. First, while we evaluate on standard CL benchmarks, TRACE, and a domain-specific Coding $\rightarrow$ Math $\rightarrow$ Medical curriculum, future work could study even longer task streams and more diverse task orders to further characterize robustness in broader deployment scenarios. Second, our implementation applies Muon-OGD mainly to matrix-valued projection layers, which are natural targets for spectral-norm-aware updates; extending the same principle to additional parameter groups or other model components may further improve performance.
\section{Numerical Details}
\label{appendix_numerical_details}
\subsection{Task Sequence and Models}
We study continual supervised fine-tuning in a three-stage curriculum:
\textbf{Stage A (coding)} $\rightarrow$ \textbf{Stage B (math)} $\rightarrow$ \textbf{Stage C (medical)}.
Unless otherwise specified, we tested our algorithm performance on Qwen2.5 family, Gemma Family, where each stage initializes from the checkpoint produced by the previous stage.

\subsection{Metrics.}
\paragraph{Average Accuracy (AA).} 
Our primary metric for evaluating overall performance is Average Accuracy (AA). This metric captures the model's final proficiency across all tasks once the entire continual learning sequence is complete. Formally, let $a_{T,i}$ denote the test accuracy on task $i$ after the model has finished training on the final task $T$. The Average Accuracy is defined as:
\begin{equation}
    \mathrm{AA} = \frac{1}{T} \sum_{i=1}^{T} a_{T,i}.
\end{equation}

\paragraph{Backward Transfer (BT).} 
To explicitly quantify the extent of catastrophic forgetting, we report Backward Transfer (BT). This metric measures the influence that learning subsequent tasks has on the retention of previously acquired skills. Let $a_{i,i}$ denote the test accuracy on task $i$ immediately after it is initially learned. The Backward Transfer after training on all $T$ tasks is calculated as:
\begin{equation}
    \mathrm{BT} = \frac{1}{T-1} \sum_{i=1}^{T-1} (a_{T,i} - a_{i,i}).
\end{equation}
A negative BT indicates that the model has forgotten prior knowledge during the sequential training stages. Conversely, a BT close to zero, or a positive value, implies that prior capabilities have been successfully protected or even positively reinforced.

\subsection{Dataset Details}
\label{app:dataset_details}

We evaluate Muon-OGD on both established continual-learning benchmarks and a domain-specific sequential instruction-tuning curriculum. The established benchmarks include the standard 15-Task Continual Learning Benchmark and TRACE, which measure long-horizon retention under heterogeneous task streams. In addition, we construct a three-stage domain sequence consisting of coding, mathematical reasoning, and medical reasoning to evaluate catastrophic forgetting in practical LLM adaptation scenarios.

\paragraph{15-Task Continual Learning Benchmark.}
The 15-Task Continual Learning Benchmark is a widely used long-sequence benchmark for evaluating catastrophic forgetting in language models. It extends the standard five-task classification benchmark into a more diverse 15-task stream. Following prior work~\cite{razdaibiedina2023progressive}, the benchmark includes five text classification datasets, AG News, Amazon Reviews, Yelp Reviews, DBpedia, and Yahoo Answers; four GLUE datasets, MNLI, QQP, RTE, and SST-2~\cite{wang2018glue}; five SuperGLUE datasets, WiC, CB, COPA, MultiRC, and BoolQ~\cite{wang2019superglue}; and the IMDB sentiment classification dataset. These tasks cover topic classification, sentiment analysis, natural language inference, paraphrase detection, semantic similarity, commonsense reasoning, reading comprehension, and yes/no question answering.

This benchmark is useful for continual learning because it evaluates both short-term adaptation and long-term retention over a long heterogeneous task sequence. Early tasks mainly require classification and topic recognition, while later tasks involve semantic reasoning and reading comprehension. In our experiments, the model learns each task sequentially without access to future-task data. After training on the full task stream, we report average accuracy (AA) across all tasks and backward transfer (BT) to quantify forgetting on previously learned tasks.

\paragraph{TRACE Benchmark.}
TRACE is a continual-learning benchmark designed for instruction-tuned large language models~\cite{wang2023trace}. Unlike traditional classification-focused continual-learning benchmarks, TRACE converts diverse datasets into a unified instruction-following format and evaluates whether an LLM can retain general capabilities while sequentially adapting to new tasks. The benchmark contains eight tasks spanning domain-specific knowledge, multilingual understanding, arithmetic reasoning, code generation, and general instruction-following ability.

TRACE provides a more realistic evaluation setting for LLM continual learning because sequential instruction tuning may degrade broad model capabilities, not only classification accuracy. For example, adapting the model to a new domain may harm its previous ability to solve mathematical problems, generate code, or follow multilingual instructions. Therefore, TRACE directly tests the stability-plasticity trade-off in aligned LLMs. In our experiments, we follow the TRACE evaluation protocol and report average accuracy (AA) and backward transfer (BT) using the LLaMA-2-7B-Chat backbone. We compare Muon-OGD with sequential fine-tuning and representative orthogonal-gradient baselines, including O-LoRA and Sculpting Subspaces.

\paragraph{Coding Dataset: BigCodeBench.}
For the coding stage in our domain-specific continual-learning sequence, we use BigCodeBench~\cite{zhuo2024bigcodebench}. BigCodeBench is a code-generation benchmark designed to evaluate practical programming ability beyond simple function-completion problems. Compared with earlier coding benchmarks such as HumanEval, BigCodeBench contains more realistic programming tasks that require understanding natural-language specifications, using external Python libraries, composing multiple operations, and generating executable code.

In our experiments, BigCodeBench is used as the first stage of the coding--mathematics--medical continual-learning sequence. We use the instruction-style setting for supervised fine-tuning and evaluation. To construct the training split, we deterministically sample 800 coding problems for supervised fine-tuning and reserve the remaining problems for held-out evaluation. Coding performance is evaluated using pass@1 under the official BigCodeBench execution-based protocol, where generated programs are executed against hidden unit tests. This stage tests whether the model can acquire executable code-generation ability and whether this ability is retained after later mathematical and medical adaptation.

\paragraph{Mathematics Dataset: GSM8K.}
For the mathematical reasoning stage, we use GSM8K~\cite{cobbe2021training}, a benchmark of grade-school math word problems. Each problem requires multi-step arithmetic reasoning and produces a final numerical answer. GSM8K is widely used for evaluating reasoning ability in language models because correct answers usually require decomposing a natural-language problem into intermediate reasoning steps rather than relying on surface-level pattern matching.

In our continual-learning sequence, GSM8K is used after the coding stage. The model is trained on the GSM8K training split, with optional subsampling controlled by the experimental configuration. Evaluation is performed on the official test split. We use exact-match numerical accuracy as the main metric after extracting the final answer from the model response. This stage evaluates whether the model can learn structured mathematical reasoning while preserving coding ability acquired in the previous stage.

\paragraph{Medical Reasoning Dataset: HuatuoGPT-o1.}
For the medical reasoning stage, we use HuatuoGPT-o1~\cite{chen2412huatuogpt}. HuatuoGPT-o1 provides medical reasoning data for supervised fine-tuning and open-ended verifiable medical questions for evaluation. This dataset is substantially different from the coding and mathematics datasets because it requires domain-specific biomedical knowledge, careful reasoning, and medically appropriate answer generation.

In training, we use the supervised fine-tuning data from \texttt{FreedomIntelligence/medical-o1-reasoning-SFT}. Our preprocessing pipeline applies an English-only filter and uses answer-span supervision, so that the model is optimized on the final medically relevant answer. For evaluation, we use the verifiable-problem set from \texttt{FreedomIntelligence/medical-o1-verifiable-problem} and sample 1,000 English questions. Since many medical answers are open-ended and cannot be reliably evaluated by exact string matching, we report verifier-based \texttt{judge\_accuracy} using a dedicated medical verifier model. This final stage is intentionally knowledge-intensive and serves as a strong test of catastrophic forgetting: after adapting to medical reasoning, the model is re-evaluated on the earlier coding and mathematics tasks to measure retention.

\paragraph{Three-Stage Domain Continual-Learning Sequence.}
In addition to the standard 15-Task and TRACE benchmarks, we evaluate Muon-OGD on a practical three-stage sequence: Coding $\rightarrow$ Mathematics $\rightarrow$ Medical. This sequence is designed to reflect realistic LLM adaptation, where a model may be sequentially specialized to different domains with very different output formats and reasoning requirements. The coding stage emphasizes executable program synthesis, the mathematics stage emphasizes symbolic and arithmetic reasoning, and the medical stage emphasizes knowledge-intensive open-ended reasoning. This setting allows us to evaluate both plasticity on the current domain and retention of previous capabilities after each stage.

\subsection{Training Details}
For the coding domain, the model is trained on the predefined 800-task training split of BigCodeBench. For mathematical reasoning, the model is trained on the GSM8K training set. Depending on the experimental setting, this data is optionally subsampled, bounded by the \texttt{num\_train\_examples} parameter. For the medical domain, we train on the HuatuoGPT-o1 SFT set. Our training pipeline applies an English-only filter, yielding approximately 20,000 training examples. We enforce answer-span supervision and explicitly remove any examples lacking supervised tokens to prevent non-finite loss issues during optimization.

\subsection{Evaluation Details}
We assess model plasticity and retention using domain-specific protocols and metrics. \textbf{Coding:} We adhere to the official BigCodeBench remote evaluation protocol. Specifically, we generate a single model completion per task (the pass@1 setting) and format the predictions into a JSON file. These are submitted to the official evaluator endpoint for execution-based scoring, and we report the resulting pass@1 accuracy.
    \textbf{Mathematics:} Performance on GSM8K is evaluated on the official \texttt{test} split, with the primary metric being exact-match numeric accuracy.
    \textbf{Medical Reasoning:} From the HuatuoGPT-o1 verifiable problem set, we randomly sample 1,000 questions for evaluation. To reliably evaluate the complex reasoning traces, we employ a dedicated verifier model (\texttt{medical\_o1\_verifier\_3B}). The primary evaluation metric is the verifier-based \texttt{judge\_accuracy}.


\begin{figure}[htbp]
\centering
\begin{tcolorbox}[
  colback=gray!8,          
  colframe=darkgray!90,    
  boxrule=1.5pt,           
  arc=5pt,                 
  fontupper=\ttfamily\small, 
  left=8pt, right=8pt, top=8pt, bottom=8pt 
]
You are a helpful math assistant. Provide the final numeric answer on the LAST line in the exact form: \#\#\#\# <number>

Solve the following problem. Output the final numeric answer on a new line as: \#\#\#\# <number>

[Math Question]\\
\{question\}
\end{tcolorbox}
\caption{Prompt template used for zero-shot mathematical evaluation on the GSM8K dataset.}
\label{fig:math_prompt}
\end{figure}

\begin{figure}[htbp]
\centering
\begin{tcolorbox}[
  colback=gray!8,          
  colframe=darkgray!90,    
  boxrule=1.5pt,           
  arc=5pt,                 
  fontupper=\ttfamily\small, 
  left=8pt, right=8pt, top=8pt, bottom=8pt 
]
You are an expert Python programmer. Please solve the following task.\\
You may briefly explain your logic, but you MUST provide your final solution inside a single Markdown code block (```python ... ```).

Here is the task:\\
\{task\_prompt\}
\end{tcolorbox}
\caption{Prompt template used for zero-shot coding evaluation on the BigCodeBench dataset.}
\label{fig:coding_prompt}
\end{figure}

\begin{figure}[htbp]
\centering
\begin{tcolorbox}[
  colback=gray!8,          
  colframe=darkgray!90,    
  boxrule=1.5pt,           
  arc=5pt,                 
  fontupper=\ttfamily\small, 
  left=8pt, right=8pt, top=8pt, bottom=8pt 
]
You are an expert medical evaluator assessing whether a model's response correctly answers a medical question. Your task is to compare the model's response to the reference answer and determine if the model's response is:\newline
1. CORRECT: The response contains the key medical information from the reference answer, even if phrased differently or includes additional correct medical details.\newline
2. INCORRECT: The response is medically wrong, misses the main point, or provides incorrect medical information.\newline
Focus on medical accuracy and completeness, not on writing style or verbosity.

[Medical Question]\newline
\{question\}\newline
[Reference Answer]\newline
\{reference\_answer\}\newline
[Model Response]\newline
\{model\_response\}\newline
Evaluate the model's response. Output ONLY one of: "CORRECT" or "INCORRECT".
\end{tcolorbox}
\caption{Prompt template used for LLM-as-a-judge evaluation on the HuatuoGPT-o1 medical dataset.}
\label{fig:medical_prompt_final}
\end{figure}

\subsection{Optimization and Hyperparameters}
\paragraph{Stage A (coding script defaults).}
\texttt{max\_length=2048}, \texttt{batch\_size=4}, \texttt{grad\_accum=4}, \texttt{epochs=3},
\texttt{lr=5e-6}, \texttt{weight\_decay=0.01}, \texttt{warmup\_ratio=0.03}, \texttt{max\_steps=500},
\texttt{max\_grad\_norm=1.0}, seed 42.

\paragraph{Stage B (math script defaults).}
\texttt{max\_length=512}, \texttt{batch\_size=1}, \texttt{grad\_accum=8}, \texttt{epochs=3},
\texttt{lr=1e-5}, \texttt{weight\_decay=0.01}, \texttt{warmup\_ratio=0.03}, \texttt{max\_steps=1800},
\texttt{max\_grad\_norm=1.0}, seed 42.

\paragraph{Stage C (medical script defaults ).}
Script defaults are \texttt{max\_length=1536}, \texttt{batch\_size=1}, \texttt{grad\_accum=8},
\texttt{epochs=1}, \texttt{lr=2e-5}, \texttt{max\_steps=1200}. For stronger adaptation, we use extended budget runs (e.g., \texttt{epochs=3}, \texttt{max\_steps=-1}, tuned warmup/LR).

\paragraph{Muon-OGD settings.}
Across scripts, Muon-OGD is enabled with constrained updates on selected projection layers (e.g., \texttt{q\_proj,k\_proj,v\_proj,o\_proj}), with key knobs:
\texttt{muon\_k}, \texttt{muon\_T}, \texttt{muon\_eta}, \texttt{muon\_eta\_dual}, matrix-sign method (\texttt{ns} or \texttt{svd}), and optional warm-start of dual variables.

\subsection{Extended Main Results}

In our extended numerical study, we evaluate the generalizability of Muon-OGD across additional architectures and scales, specifically examining the Qwen2.5-7B, LLaMA3.2-1B-instruct, and LLaMA3.2-3B-instruct models on the domain-specific Coding $\rightarrow$ Math $\rightarrow$ Medical sequence. The results, detailed in Tables \ref{tab:alt_sequence_results_7b}, \ref{tab:alt_sequence_results_llama1b}, and \ref{tab:alt_sequence_results_llama3b}, confirm that the advantages of spectral-norm-aware orthogonal projection extend reliably across different LLM backbones.

\paragraph{Performance on the LLaMA3.2 Family.}
Tables \ref{tab:alt_sequence_results_llama1b} and \ref{tab:alt_sequence_results_llama3b} demonstrate that Muon-OGD consistently outperforms the baselines on both the 1B and 3B LLaMA3.2 architectures. For the 1B model, Muon-OGD achieves the highest Average Accuracy (AA) at every stage, culminating in a final Stage C AA of $26.2\%$, outperforming SeqSFT ($24.2\%$) and Sculpting Subspaces ($23.9\%$). The advantage is even more pronounced in the 3B model, where Muon-OGD sweeps all stages. Notably, Muon-OGD exhibits exceptional knowledge retention: after training on the final Medical domain, the 3B model retains a Math accuracy of $55.1\%$, suffering minimal catastrophic forgetting compared to its peak Stage B Math score of $56.0\%$. In contrast, baselines like O-LoRA and SeqSFT experience sharper drops in previously acquired skills.

\paragraph{Scaling to Qwen2.5-7B.}
Table \ref{tab:alt_sequence_results_7b} extends our evaluation to the larger Qwen2.5-7B-Instruct model, incorporating Gradient Replay to further anchor prior knowledge. While O-LoRA shows strong competitive performance at Stage B (edging out Muon-OGD in AA by $0.8\%$), Muon-OGD dominates the final curriculum evaluation at Stage C. At the conclusion of the three-stage sequence, Muon-OGD achieves the highest accuracy across all three individual domains---Coding ($38.1\%$), Math ($75.2\%$), and Medical ($35.6\%$)---yielding the highest overall AA of $49.6\%$. 

Overall, these extended results reinforce our primary findings. Integrating the Muon optimizer's operator-norm geometry with projection-based constraints provides a highly robust mechanism for balancing plasticity on new domains while safeguarding previously acquired reasoning capabilities.

\begin{table}[htbp]
\centering
\small
\setlength{\tabcolsep}{3.5pt} 
\caption{Continual learning performance of the Qwen2.5-7B-instruct architecture evaluated across a sequential three-stage curriculum (Coding $\rightarrow$ Math $\rightarrow$ Medical). Muon-OGD utilizes Gradient Replay to prevent catastrophic forgetting. All metrics represent testing accuracy percentages (\%). For each metric within a specific evaluation stage, the best performance is highlighted in \textbf{bold}, and the second-best is \underline{underlined}.}
\label{tab:alt_sequence_results_7b}
\begin{tabular}{@{} llcccc @{}} 
\toprule
\textbf{Stage} & \textbf{Optimizer} & \textbf{Coding} & \textbf{Math} & \textbf{Medical} & \textbf{AA} \\
\midrule

\multirow{4}{*}{\textbf{A (Coding)}} 
& SeqSFT & 37.8 & - & - & 37.8 \\
& O-LoRA & 37.9 & - & - & 37.9 \\
& Sculpting Sub. & \underline{38.8} & - & - & \underline{38.8} \\
& Muon-OGD & \textbf{40.8} & - & - & \textbf{40.8} \\

\midrule

\multirow{4}{*}{\textbf{B (Math)}} 
& SeqSFT & \textbf{38.8} & 72.0 & - & 55.4 \\
& O-LoRA & \textbf{38.8} & \textbf{75.0} & - & \textbf{56.9} \\
& Sculpting Sub. & \underline{38.6} & 72.4 & - & 55.5 \\
& Muon-OGD & 38.2 & \underline{74.0} & - & \underline{56.1} \\

\midrule

\multirow{4}{*}{\textbf{C (Medical)}} 
& SeqSFT & 36.1 & 71.0 & 34.4 & 47.2 \\
& O-LoRA & 37.5 & \underline{74.0} & 35.0 & \underline{48.8} \\
& Sculpting Sub. & \underline{37.9} & 73.2 & \underline{35.2} & \underline{48.8} \\
& Muon-OGD & \textbf{38.1} & \textbf{75.2} & \textbf{35.6} & \textbf{49.6} \\

\bottomrule
\end{tabular}
\end{table}


\begin{table}[htbp]
\centering
\small
\setlength{\tabcolsep}{3.5pt}
\caption{Comparative performance of LLaMA3.2-1B-instruct across sequential training stages (Coding $\rightarrow$ Math $\rightarrow$ Medical). Average Accuracy (AA) measures the mean performance across all three tasks at each stage. Bold values indicate the best performance, while underlined values denote the runner-up.}
\label{tab:alt_sequence_results_llama1b}
\begin{tabular}{@{} llcccc @{}}
\toprule
\textbf{Stage} & \textbf{Optimizer} & \textbf{Coding} & \textbf{Math} & \textbf{Medical} & \textbf{AA} \\

\midrule

\multirow{4}{*}{\textbf{A (Coding)}} 
& SeqSFT         & 16.7 & - & - & 16.7 \\
& O-LoRA         & 14.7 & - & - & 14.7 \\
& Sculpting Sub. & \underline{18.7} & - & - & \underline{18.7} \\
& Muon-OGD       & \textbf{20.8}$_{\pm 1.1}$ & - & - & \textbf{20.8}$_{\pm 1.1}$ \\

\midrule

\multirow{4}{*}{\textbf{B (Math)}} 
& SeqSFT         & \textbf{21.7} & 34.6 & - & \underline{28.2} \\
& O-LoRA         & 18.7 & \textbf{35.8} & - & 27.3 \\
& Sculpting Sub. & 20.7 & 35.6 & - & \underline{28.2} \\
& Muon-OGD       & \textbf{21.7}$_{\pm 1.2}$ & \underline{35.6}$_{\pm 1.4}$ & - & \textbf{28.7}$_{\pm 0.9}$ \\

\midrule

\multirow{4}{*}{\textbf{C (Medical)}} 
& SeqSFT         & \underline{16.7} & \underline{33.8} & \underline{22.2} & \underline{24.2} \\
& O-LoRA         & 11.7 & 31.6 & 22.0 & 21.8 \\
& Sculpting Sub. & \underline{17.6} & 32.2 & 22.0 & 23.9 \\
& Muon-OGD       & \textbf{19.7}$_{\pm 0.8}$ & \textbf{35.5}$_{\pm 1.4}$ & \textbf{23.5}$_{\pm 0.7}$ & \textbf{26.2}$_{\pm 0.6}$ \\

\bottomrule
\end{tabular}
\end{table}


\begin{table}[htbp]
\centering
\small
\setlength{\tabcolsep}{3.5pt}
\caption{Continual learning performance of the LLaMA3.2-3B-instruct architecture evaluated across a sequential three-stage curriculum (Coding $\rightarrow$ Math $\rightarrow$ Medical). All metrics represent testing accuracy percentages (\%). For each metric within a specific evaluation stage, the best performance is highlighted in \textbf{bold}, and the second-best is \underline{underlined}.}
\label{tab:alt_sequence_results_llama3b}
\begin{tabular}{@{} llcccc @{}}
\toprule
\textbf{Stage} & \textbf{Optimizer} & \textbf{Coding} & \textbf{Math} & \textbf{Medical} & \textbf{AA} \\
& & (800 Q) & (GSM8K) & (Huatuo) & \\

\midrule

\multirow{4}{*}{\textbf{A (Coding)}} 
& SeqSFT         & 24.4 & - & - & 24.4 \\
& O-LoRA         & \underline{25.8} & - & - & \underline{25.8} \\
& Sculpting Sub. & 24.7 & - & - & 24.7 \\
& Muon-OGD       & \textbf{28.3}$_{\pm 1.6}$ & - & - & \textbf{28.3}$_{\pm 1.6}$ \\

\midrule

\multirow{4}{*}{\textbf{B (Math)}} 
& SeqSFT         & 20.0 & 53.0 & - & 36.5 \\
& O-LoRA         & 21.4 & \textbf{55.0} & - & 38.2 \\
& Sculpting Sub. & \underline{23.2} & \textbf{55.0} & - & \underline{39.1} \\
& Muon-OGD       & \textbf{26.4}$_{\pm 0.8}$ & \textbf{56.0}$_{\pm 1.9}$ & - & \textbf{41.2}$_{\pm 1.0}$ \\

\midrule

\multirow{4}{*}{\textbf{C (Medical)}} 
& SeqSFT         & 17.0 & 50.0 & \underline{31.8} & 32.9 \\
& O-LoRA         & 18.2 & 52.4 & \underline{31.8} & 34.1 \\
& Sculpting Sub. & \underline{20.1} & \underline{53.8} & \underline{31.8} & \underline{35.2} \\
& Muon-OGD       & \textbf{25.1}$_{\pm 0.5}$ & \textbf{55.1}$_{\pm 1.5}$ & \textbf{32.9}$_{\pm 1.2}$ & \textbf{37.7}$_{\pm 0.7}$ \\

\bottomrule
\end{tabular}
\end{table}
\section{Extended Numerical Study}
\label{sec:ablation}

To comprehensively understand the sensitivity and contribution of the core hyperparameters in the Muon-OGD framework, we conduct an extensive ablation study. Because our evaluation spans diverse domains, we analyze these factors across two distinct settings: (1) the standard and 15-task NLP benchmarks using the encoder-decoder T5-Large, and (2) the highly specialized Coding $\rightarrow$ Math $\rightarrow$ Medical sequence using the decoder-only Qwen2.5-1.5B-instruct architecture. Unless otherwise stated, the primary evaluation metric is the final Average Accuracy (AA).

\subsection{Learning Rate Allocation and Plasticity}

Muon-OGD relies on a dual-branch update mechanism, necessitating a careful balance between the matrix-level plasticity (controlled by the Muon step size) and the scalar/vector updates (controlled by the AdamW base learning rate).

\paragraph{Muon / Primal Step Size (\texttt{MUON\_LR} / \texttt{muon\_eta}).} 
The Muon step size directly controls the matrix-level plasticity for the targeted 2D parameters (e.g., $q, k, v, o$). Table~\ref{tab:ablation_muon_lr} shows that increasing \texttt{MUON\_LR} substantially improves performance on the T5 standard benchmark, raising the average accuracy from 74.3\% to 77.7\%. On the 15-task benchmark, $5\times10^{-3}$ emerges as the optimal value. Conversely, for the Qwen2.5 architecture (Figure~\ref{fig:ablation_avg}), the scaling factor for the orthogonalized momentum (\texttt{muon\_eta}) exhibits a slow degradation if overly aggressive. A conservative value of $10^{-6}$ provides the highest stability for LLMs, ensuring unconstrained updates do not distort the established geometry.

\begin{table}[htbp]
\centering
\small
\setlength{\tabcolsep}{7pt}
\renewcommand{\arraystretch}{1.08}
\caption{Effect of Muon step size. For the standard benchmark, we report the average over Orders 1--3. For the large benchmark, we report Large Order-4. All other hyperparameters are fixed within each block.}
\label{tab:ablation_muon_lr}
\begin{tabular}{lcc}
\toprule
\textbf{Setting} & \textbf{\texttt{MUON\_LR}} & \textbf{AA (\%)} \\
\midrule
\multicolumn{3}{l}{\textit{Standard benchmark}} \\
Standard & $1{\times}10^{-3}$ & 74.3 \\
Standard & $2{\times}10^{-3}$ & 76.5 \\
Standard & $5{\times}10^{-3}$ & \textbf{77.7} \\
\midrule
\multicolumn{3}{l}{\textit{15-task benchmark, Order-4}} \\
Large & $4{\times}10^{-3}$ & 67.8 \\
Large & $5{\times}10^{-3}$ & \textbf{72.2} \\
Large & $6{\times}10^{-3}$ & 70.5 \\
\bottomrule
\end{tabular}
\end{table}

\paragraph{AdamW Base Learning Rate (\texttt{LR} / \texttt{lr}).} 
Reducing the AdamW branch learning rate is not a viable substitute for explicit subspace protection. As shown in Table~\ref{tab:ablation_adam_lr}, suppressing \texttt{LR} to $3\times10^{-5}$ on T5 causes severe underfitting on the current task (e.g., Amazon collapses). A similar high sensitivity is observed in Qwen2.5 (Figure~\ref{fig:ablation_avg}), where AA monotonically increases as \texttt{lr} approaches $5 \times 10^{-5}$. This confirms that Muon-OGD requires sufficient baseline plasticity combined with strictly enforced orthogonal corrections.

\begin{table}[htbp]
\centering
\small
\setlength{\tabcolsep}{7pt}
\renewcommand{\arraystretch}{1.08}
\caption{Effect of the AdamW-branch learning rate on the standard benchmark (\texttt{MUON\_LR}$=1{\times}10^{-3}$).}
\label{tab:ablation_adam_lr}
\begin{tabular}{ccp{5.2cm}}
\toprule
\textbf{\texttt{LR}} & \textbf{AA (\%)} & \textbf{Observation} \\
\midrule
$3{\times}10^{-5}$ & 62.4 & Too small; Amazon collapses to near-random performance. \\
$3{\times}10^{-4}$ & 71.0 & Better, but still below the default setting. \\
$1{\times}10^{-3}$ & \textbf{75.7} & Sufficient current-task plasticity. \\
$2{\times}10^{-3}$ & 74.5 & Increasing LR further does not improve performance. \\
\bottomrule
\end{tabular}
\end{table}

\subsection{Dual Correction Strength (\texttt{MUON\_ETA\_DUAL})}

The dual correction dictates how aggressively the optimizer enforces the protected-subspace constraint ($U^\top \Delta V \approx 0$). Table~\ref{tab:ablation_dual_eta} demonstrates that this parameter has a strict sweet spot for long task sequences. On the 15-task T5 benchmark, a weak step ($1\times10^{-4}$) fails to protect old-task directions, while a strong step ($5\times10^{-4}$) over-constrains new-task plasticity. The intermediate $3\times10^{-4}$ performs best. This behavior is mirrored perfectly in the Qwen2.5 experiments (Figure~\ref{fig:ablation_avg}), where a distinct optimal peak is observed at $5 \times 10^{-5}$, representing the critical threshold for balancing constraint satisfaction with model plasticity.

\begin{table}[htbp]
\centering
\small
\setlength{\tabcolsep}{8pt}
\renewcommand{\arraystretch}{1.08}
\caption{Effect of dual correction strength on the 15-task benchmark (Large Order-4).}
\label{tab:ablation_dual_eta}
\begin{tabular}{cc}
\toprule
\textbf{\texttt{MUON\_ETA\_DUAL}} & \textbf{AA (\%)} \\
\midrule
$1{\times}10^{-4}$ & 67.3 \\
$3{\times}10^{-4}$ & \textbf{72.2} \\
$5{\times}10^{-4}$ & 66.5 \\
\bottomrule
\end{tabular}
\end{table}

\subsection{Protected Subspace Capacity and Rank (\texttt{muon\_k})}

Table~\ref{tab:rank_capacity_exploratory} and Figure~\ref{fig:ablation_avg} explore the allocation of protected rank ($k$). Both architectures reveal that protected-rank allocation is highly sensitive but non-monotonic. For T5, allocating a moderate protected rank ($0.08/\mathrm{max}48$) yields the best trade-off, indicating the subspace must be large enough to retain structure but compact enough to permit future learning. For Qwen2.5, performance remains remarkably robust across a wide range ($k=1$ to $20$), with a localized optimal peak at $k=3$. This suggests that for highly specialized reasoning tasks, critical prior knowledge is heavily concentrated in a top-dominant, compact singular subspace.

\begin{table}[htbp]
\centering
\small
\setlength{\tabcolsep}{7pt}
\renewcommand{\arraystretch}{1.08}
\caption{Exploratory analysis of protected-rank allocation on T5 Large Order-4.}
\label{tab:rank_capacity_exploratory}
\begin{tabular}{cc}
\toprule
\textbf{Protected rank / capacity} & \textbf{AA (\%)} \\
\midrule
$0.10 / \mathrm{max}64 / \mathrm{total}768$ & 68.1 \\
$0.05 / \mathrm{max}32 / \mathrm{total}1024$ & 64.6 \\
$0.08 / \mathrm{max}48 / \mathrm{total}1024$ & \textbf{72.2} \\
\bottomrule
\end{tabular}
\end{table}

\subsection{Orthogonalization Precision}

\begin{figure}[htbp]
  \centering
  \makebox[\textwidth][c]{\includegraphics[width=1.0\textwidth]{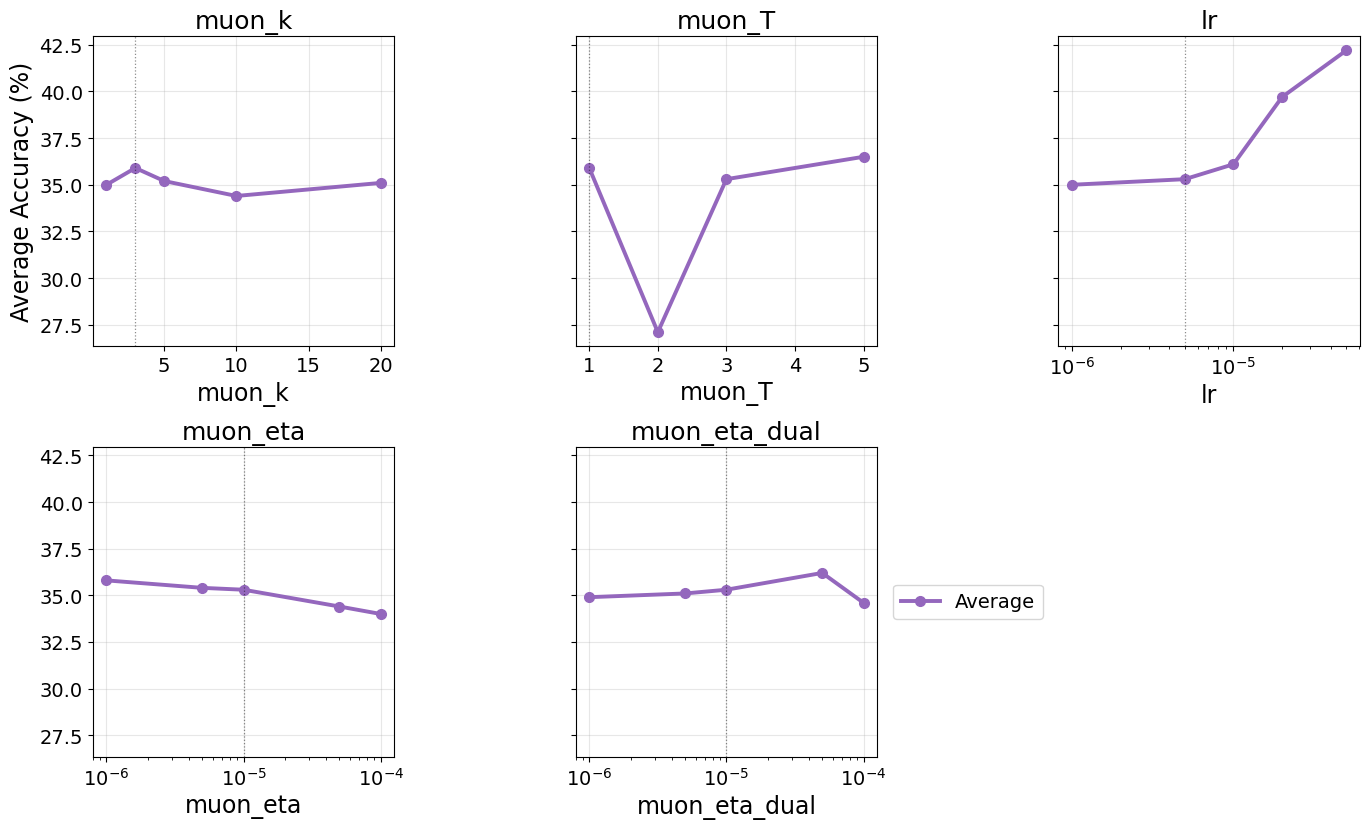}}
  \caption{Ablation study of Muon-OGD hyperparameters on the Qwen2.5-1.5B architecture. The figure illustrates the impact of varying the protected subspace rank (\texttt{muon\_k}), Newton-Schulz iterations (\texttt{muon\_T}), base learning rate (\texttt{lr}), primal step size (\texttt{muon\_eta}), and dual step size (\texttt{muon\_eta\_dual}) on the final average accuracy.}
  \label{fig:ablation_avg}
\end{figure}

The computational precision of the orthogonal projections is controlled by Randomized SVD iterations (\texttt{SVD\_NITER}) in T5, and Newton-Schulz iterations (\texttt{muon\_T}) in Qwen2.5. Table~\ref{tab:appendix_svd_niter} shows that increasing \texttt{SVD\_NITER} from 2 to 4 yields marginal gains. Similarly, Figure~\ref{fig:ablation_avg} highlights a non-linear response for \texttt{muon\_T}, with strong performance at $T=1$ and a dip at $T=2$. Given the $\mathcal{O}(N^3)$ overhead of iterative matrix multiplications, lower iteration counts ($T=1$, \texttt{SVD\_NITER}=4) provide the optimal empirical balance between projection accuracy and step-time efficiency.

\begin{table}[htbp]
\centering
\small
\setlength{\tabcolsep}{7pt}
\renewcommand{\arraystretch}{1.08}
\caption{Effect of randomized SVD iterations on the T5 standard benchmark. Fixed hyperparameters: LR=$1\times10^{-3}$, MUON\_LR=$1\times10^{-3}$, MUON\_ETA\_DUAL=$1\times10^{-4}$.}
\label{tab:appendix_svd_niter}
\begin{tabular}{ccccc}
\toprule
\textbf{\texttt{SVD\_NITER}} & \textbf{Order-1} & \textbf{Order-2} & \textbf{Order-3} & \textbf{Avg} \\
\midrule
2 & 75.7 & 72.9 & 74.2 & 74.3 \\
4 & 75.5 & 73.7 & 73.6 & 74.3 \\
\bottomrule
\end{tabular}
\end{table}

\subsection{Summary of Final Configurations}

Table~\ref{tab:final_hyperparams} outlines the final tuned parameters for the T5 NLP benchmarks. For the Qwen2.5 benchmarks, we established the default configuration as $\texttt{lr} = 5 \times 10^{-5}$, $\texttt{muon\_T} = 1$, $\texttt{muon\_eta\_dual} = 5 \times 10^{-5}$, $\texttt{muon\_eta} = 10^{-6}$, and $\texttt{muon\_k} = 3$. 

\subsection{Experimental setup and computational resources.}
We evaluate Muon-OGD on two families of continual learning benchmarks: (i) the standard continual learning benchmarks following the O-LoRA protocol~\citep{wang2023orthogonal} \textcolor{blue}, and (ii) the TRACE aligned-LLM continual learning benchmark~\citep{wang2023trace}.  For all continual learning experiments—including the 5-task and 15-task benchmarks, as well as the TRACE benchmark—we replicate the task sequences, prompts, and dataset configurations as established in the original works.

For the standard continual learning benchmarks, we use the encoder-decoder T5-Large model as the backbone and follow the O-LoRA benchmark protocol.  The standard 5-task benchmark contains task Orders 1--3, while the 15-task long-sequence benchmark contains Orders 4--6.  All Muon-OGD runs for T5-Large are executed on a single H100 80GB GPU using standard PyTorch training with bf16 precision and gradient checkpointing.  We use a per-device batch size of 8 with gradient accumulation of 8, resulting in an effective batch size of 64, matching the O-LoRA T5 training protocol. For the TRACE benchmark, we use LLaMA-2-7B-Chat as the backbone. All experiments are conducted on a server equipped with 8 NVIDIA H100 GPUs using DeepSpeed ZeRO Stage 2 with gradient checkpointing enabled. We use a per-GPU micro-batch size of 1 with gradient accumulation steps of 1, resulting in an effective training batch size of 8.

Unless otherwise specified, the main reported metric is final-stage average accuracy (AA), computed by evaluating all tasks in the stream after training on the final task. For both T5 and TRACE benchmarks, we report the mean over three independent random seeds unless otherwise noted.For T5 evaluation, we use a maximum source length of 512 and a maximum generation length of 50. For TRACE evaluation, we use a maximum prompt length of 1024 and a maximum answer length of 512. The full task sequences are listed in Table~\ref{tab:task_orders_all}, and the main training hyperparameters are summarized in Table~\ref{tab:final_hyperparams}.

\begin{table}[htbp]
\centering
\small
\setlength{\tabcolsep}{5pt}
\renewcommand{\arraystretch}{1.08}
\caption{Final hyperparameter choices for Muon-OGD on the Standard and TRACE continual learning benchmarks.}
\label{tab:final_hyperparams}
\begin{tabular}{lccc}
\toprule
\textbf{Hyperparameter} 
& \textbf{Standard Benchmark} 
& \textbf{15-Task Benchmark} 
& \textbf{TRACE} \\
\midrule
\texttt{LR} 
& $1{\times}10^{-3}$ 
& $1{\times}10^{-3}$ 
& $1{\times}10^{-5}$ \\

\texttt{MUON\_LR} 
& $5{\times}10^{-3}$ 
& $5{\times}10^{-3}$ 
& $1{\times}10^{-5}$ \\

\texttt{MUON\_T} 
& 1 
& 2 
& 1 \\

\texttt{MUON\_ETA\_DUAL} 
& $1{\times}10^{-4}$ 
& $3{\times}10^{-4}$ 
& $1{\times}10^{-5}$ \\

Max protected rank 
& 64 
& 48 
& 32 \\

\texttt{SVD\_NITER} 
& 4 
& 4 
& 4 \\
\bottomrule
\end{tabular}
\vspace{2pt}
\end{table}

\begin{table*}[htbp]
\centering
\setlength{\tabcolsep}{8pt} 
\renewcommand{\arraystretch}{1.3} 
\caption{
Task orders used in the continual learning benchmarks.
Orders 1--3 correspond to the standard benchmark, Orders 4--6 correspond to the 15-task long-sequence benchmark, and the TRACE order corresponds to the aligned-LLM continual learning benchmark.
}
\label{tab:task_orders_all}
\begin{tabular}{@{} c l p{0.65\textwidth} @{}} 
\toprule
\textbf{Order} & \textbf{Benchmark} & \textbf{Task sequence} \\
\midrule
Order-1 
& T5 Standard 
& DBpedia $\rightarrow$ Amazon $\rightarrow$ Yahoo $\rightarrow$ AGNews \\

Order-2 
& T5 Standard 
& DBpedia $\rightarrow$ Amazon $\rightarrow$ AGNews $\rightarrow$ Yahoo \\

Order-3 
& T5 Standard 
& Yahoo $\rightarrow$ Amazon $\rightarrow$ AGNews $\rightarrow$ DBpedia \\

\midrule
Order-4 
& T5 15-task 
& MNLI $\rightarrow$ CB $\rightarrow$ WiC $\rightarrow$ COPA $\rightarrow$ QQP $\rightarrow$ BoolQA $\rightarrow$ RTE $\rightarrow$ IMDB $\rightarrow$ Yelp $\rightarrow$ Amazon $\rightarrow$ SST-2 $\rightarrow$ DBpedia $\rightarrow$ AGNews $\rightarrow$ MultiRC $\rightarrow$ Yahoo \\

Order-5 
& T5 15-task 
& MultiRC $\rightarrow$ BoolQA $\rightarrow$ WiC $\rightarrow$ MNLI $\rightarrow$ CB $\rightarrow$ COPA $\rightarrow$ QQP $\rightarrow$ RTE $\rightarrow$ IMDB $\rightarrow$ SST-2 $\rightarrow$ DBpedia $\rightarrow$ AGNews $\rightarrow$ Yelp $\rightarrow$ Amazon $\rightarrow$ Yahoo \\

Order-6 
& T5 15-task 
& Yelp $\rightarrow$ Amazon $\rightarrow$ MNLI $\rightarrow$ CB $\rightarrow$ COPA $\rightarrow$ QQP $\rightarrow$ RTE $\rightarrow$ IMDB $\rightarrow$ SST-2 $\rightarrow$ DBpedia $\rightarrow$ AGNews $\rightarrow$ Yahoo $\rightarrow$ MultiRC $\rightarrow$ BoolQA $\rightarrow$ WiC \\

\midrule
TRACE 
& TRACE 
& C-STANCE $\rightarrow$ FOMC $\rightarrow$ MeetingBank $\rightarrow$ Py150 $\rightarrow$ ScienceQA $\rightarrow$ NumGLUE-cm $\rightarrow$ NumGLUE-ds $\rightarrow$ 20Minuten \\
\bottomrule
\end{tabular}
\end{table*}

\begin{table}[htbp]
\centering
\small
\caption{Per-task performance on the TRACE benchmark with LLaMA-2-7B-Chat.}
\label{tab:trace_llama2_muonogd_multiseed}
\begin{tabular}{lcc}
\toprule
\textbf{Task} & \textbf{After Task (\%)} & \textbf{Final (\%)} \\
\midrule
C-STANCE      & 43.4$_{\pm1.3}$ & 44.6$_{\pm1.4}$ \\
FOMC          & 71.6$_{\pm2.8}$ & 58.9$_{\pm2.1}$ \\
MeetingBank   & 52.0$_{\pm0.2}$ & 49.2$_{\pm0.9}$ \\
Py150         & 51.7$_{\pm0.8}$ & 51.3$_{\pm1.0}$ \\
ScienceQA     & 84.6$_{\pm0.0}$ & 70.0$_{\pm3.1}$ \\
NumGLUE-cm    & 27.2$_{\pm2.1}$ & 25.1$_{\pm0.7}$ \\
NumGLUE-ds    & 57.0$_{\pm0.4}$ & 55.6$_{\pm0.7}$ \\
20Minuten     & 40.7$_{\pm0.1}$ & 40.7$_{\pm0.1}$ \\
\midrule
\textbf{Average} & \textbf{53.5$_{\pm0.8}$} & \textbf{49.4$_{\pm0.3}$} \\
\bottomrule
\end{tabular}
\end{table}

\begin{table}[htbp]
\centering
\small
\caption{Order-1 per-task results (T5-Large): DBpedia $\rightarrow$ Amazon $\rightarrow$ Yahoo $\rightarrow$ AGNews.}
\label{tab:standard_order1_per_task_bt}
\begin{tabular}{lcc}
\toprule
\textbf{Task} & \textbf{After Task} & \textbf{Final} \\
\midrule
DBpedia & 98.9$_{\pm0.0}$ & 98.3$_{\pm0.4}$ \\
Amazon  & 59.9$_{\pm1.2}$ & 55.3$_{\pm3.1}$ \\
Yahoo   & 73.8$_{\pm0.3}$ & 71.6$_{\pm0.7}$ \\
AGNews  & 90.5$_{\pm0.1}$ & 90.5$_{\pm0.1}$ \\
\midrule
\multicolumn{3}{c}{\small AA: 78.9$_{\pm0.8}$ \quad BT: $-2.5_{\pm0.6}$} \\
\bottomrule
\end{tabular}
\end{table}
\begin{table}[htbp]
\centering
\small
\caption{Order-2 per-task results (T5-Large): DBpedia $\rightarrow$ Amazon $\rightarrow$ AGNews $\rightarrow$ Yahoo.}
\label{tab:standard_order2_per_task_bt}
\begin{tabular}{lcc}
\toprule
\textbf{Task} & \textbf{After Task} & \textbf{Final} \\
\midrule
DBpedia & 98.9$_{\pm0.0}$ & 98.4$_{\pm0.2}$ \\
Amazon  & 59.9$_{\pm1.2}$ & 57.7$_{\pm2.3}$ \\
AGNews  & 90.0$_{\pm0.3}$ & 87.6$_{\pm0.4}$ \\
Yahoo   & 74.0$_{\pm1.0}$ & 74.0$_{\pm1.0}$ \\
\midrule
\multicolumn{3}{c}{\small AA: 79.4$_{\pm0.7}$ \quad BT: $-1.7_{\pm0.5}$} \\
\bottomrule
\end{tabular}
\end{table}

\begin{table}[htbp]
\centering
\small
\caption{Order-3 per-task results (T5-Large): Yahoo $\rightarrow$ Amazon $\rightarrow$ AGNews $\rightarrow$ DBpedia.}
\label{tab:standard_order3_per_task_bt}
\begin{tabular}{lcc}
\toprule
\textbf{Task} & \textbf{After Task} & \textbf{Final} \\
\midrule
Yahoo   & 73.8$_{\pm0.5}$ & 70.5$_{\pm1.1}$ \\
Amazon  & 60.2$_{\pm0.5}$ & 55.7$_{\pm4.0}$ \\
AGNews  & 89.9$_{\pm0.4}$ & 87.9$_{\pm1.0}$ \\
DBpedia & 98.9$_{\pm0.1}$ & 98.9$_{\pm0.1}$ \\
\midrule
\multicolumn{3}{c}{\small AA: 78.3$_{\pm1.0}$ \quad BT: $-3.3_{\pm1.4}$} \\
\bottomrule
\end{tabular}
\end{table}

\begin{table}[htbp]
\centering
\small
\caption{Order-4 per-task results (T5-Large): MNLI $\rightarrow$ CB $\rightarrow$ WiC $\rightarrow$ COPA $\rightarrow$ QQP $\rightarrow$ BoolQA $\rightarrow$ RTE $\rightarrow$ IMDB $\rightarrow$ Yelp $\rightarrow$ Amazon $\rightarrow$ SST-2 $\rightarrow$ DBpedia $\rightarrow$ AGNews $\rightarrow$ MultiRC $\rightarrow$ Yahoo.}
\label{tab:large_order4_per_task_bt}
\begin{tabular}{lcc}
\toprule
\textbf{Task} & \textbf{After Task} & \textbf{Final} \\
\midrule
MNLI    & 84.9$_{\pm0.4}$ & 47.2$_{\pm5.2}$ \\
CB      & 89.9$_{\pm1.7}$ & 60.7$_{\pm2.9}$ \\
WiC     & 55.7$_{\pm4.3}$ & 56.3$_{\pm1.6}$ \\
COPA    & 45.7$_{\pm1.7}$ & 58.0$_{\pm4.5}$ \\
QQP     & 85.0$_{\pm2.4}$ & 79.3$_{\pm4.8}$ \\
BoolQA  & 82.8$_{\pm0.3}$ & 76.0$_{\pm2.5}$ \\
RTE     & 86.8$_{\pm1.2}$ & 75.6$_{\pm5.5}$ \\
IMDB    & 93.9$_{\pm0.3}$ & 94.2$_{\pm0.5}$ \\
Yelp    & 62.6$_{\pm2.1}$ & 52.5$_{\pm2.5}$ \\
Amazon  & 60.3$_{\pm1.0}$ & 51.7$_{\pm2.3}$ \\
SST-2   & 94.8$_{\pm0.4}$ & 93.1$_{\pm1.7}$ \\
DBpedia & 98.6$_{\pm0.4}$ & 98.7$_{\pm0.0}$ \\
AGNews  & 90.1$_{\pm0.0}$ & 89.2$_{\pm0.6}$ \\
MultiRC & 73.6$_{\pm1.6}$ & 76.0$_{\pm0.2}$ \\
Yahoo   & 74.3$_{\pm0.4}$ & 74.3$_{\pm0.4}$ \\
\midrule
\multicolumn{3}{c}{\small AA: 72.2$_{\pm0.8}$ \quad BT: $-6.9_{\pm0.9}$} \\
\bottomrule
\end{tabular}
\end{table}


\end{document}